\documentclass[11pt]{article}

\usepackage[final]{acl}
\usepackage{times}
\usepackage{latexsym}

\usepackage[T1]{fontenc}
\usepackage[utf8]{inputenc}

\usepackage{microtype}

\usepackage{inconsolata}

\usepackage{graphicx}
\usepackage{amsmath}
\usepackage{amssymb}
\usepackage{booktabs}
\usepackage{hyperref}
\title{EmoStance: Response-Side Affective-Orientation Control for Empathetic Response Generation via Emoji Weak Supervision}

\author{
  \textbf{Ziyuan Jin},
  \textbf{Yuxuan Ge},
  \textbf{Zheng Tian}\textsuperscript{\ensuremath{\dagger}} \\
  ShanghaiTech University, Shanghai, China \\
  \texttt{\{jinzy2024, geyx2023, tianzheng\}@shanghaitech.edu.cn} \\
  \small{
    \textsuperscript{\ensuremath{\dagger}}Corresponding author.
  }
}

\begin{document}
\maketitle
\begin{abstract}
Empathetic response generation requires models to decide not only what to say,
but also how to respond to the previous speaker's affective situation. We
formulate this as response-side affective-orientation control and use
multi-annotator emoji distributions as weak affective--attitudinal evidence,
rather than as output symbols or gold labels, to induce a latent control space
that operationally approximates listener stance. We construct
\textsc{EmojiDialogue}, an utterance-level extension of
\textsc{EmpatheticDialogues} with emoji votes and confidence scores, and propose
\textsc{EmoStance}, which models source-side affective expression, predicts a
soft response-side orientation from dialogue context and speaker roles, and
steers a frozen instruction-tuned LLM through continuous prefix embeddings. In
blind pairwise evaluation with 20 annotators and 800 judgments,
\textsc{EmoStance} achieves a 62.2\% decisive win rate, with the clearest gains
in contextual specificity and perceived responsiveness, while remaining
complementary to external-knowledge methods. Code, annotation metadata, and reconstruction scripts are available in our GitHub repository: \href{https://github.com/18277390221/EmoStance}{https://github.com/18277390221/EmoStance}.

\end{abstract}

\section{Introduction}

Empathetic dialogue generation requires models to decide not only what to say,
but also how the next speaker should take up the previous turn. A response can
be topically relevant yet still feel detached, overly cheerful, intrusive,
didactic, or insufficiently responsive to the speaker's affective situation. We
refer to this operational variable as \emph{response-side affective orientation}:
a soft representation of how the next reply should be affectively and
interpersonally oriented before it is verbalized. This notion is related to
\emph{listener stance}, but we do not assume access to direct or gold
listener-stance labels; instead, we treat listener stance as a higher-level
interpretation of a weakly supervised response-side control representation.

Existing supervision only partially captures this orientation. Prior work represents affective context through situation-level
emotion labels \citep{rashkin-etal-2019-towards},
dimensional affect representations
\citep{mohammad-2018-obtaining,
colombo-etal-2019-affect-driven},
and support-strategy taxonomies, such as questioning,
reflection, suggestion, and information provision
\citep{liu-etal-2021-towards}.
Subsequent systems model turn-level state transitions,
mixed initiative, strategy-response decoupling, or discourse
dynamics
\citep{zhao-etal-2023-transesc,
deng-etal-2023-knowledge,
zhang-etal-2025-decoupledesc,
wan-etal-2025-emodynamix}.
These variables characterize the dialogue state, but do not
by themselves determine whether the next response should
realize reassurance, shared excitement, gentle concern,
cautious probing, or another interpersonal orientation.
Prompt-based interfaces have also been studied for smooth
control of predefined attribute intensity
\citep{zhou-etal-2024-evaluating}.
Our target instead is a response-specific, mixture-like latent
orientation that may be difficult to verbalize as a short and
stable instruction. These limitations motivate a soft
intermediate variable for modeling how the next response
should be affectively and interpersonally positioned.

\begin{figure}[t]
    \centering
    \includegraphics[width=\columnwidth]{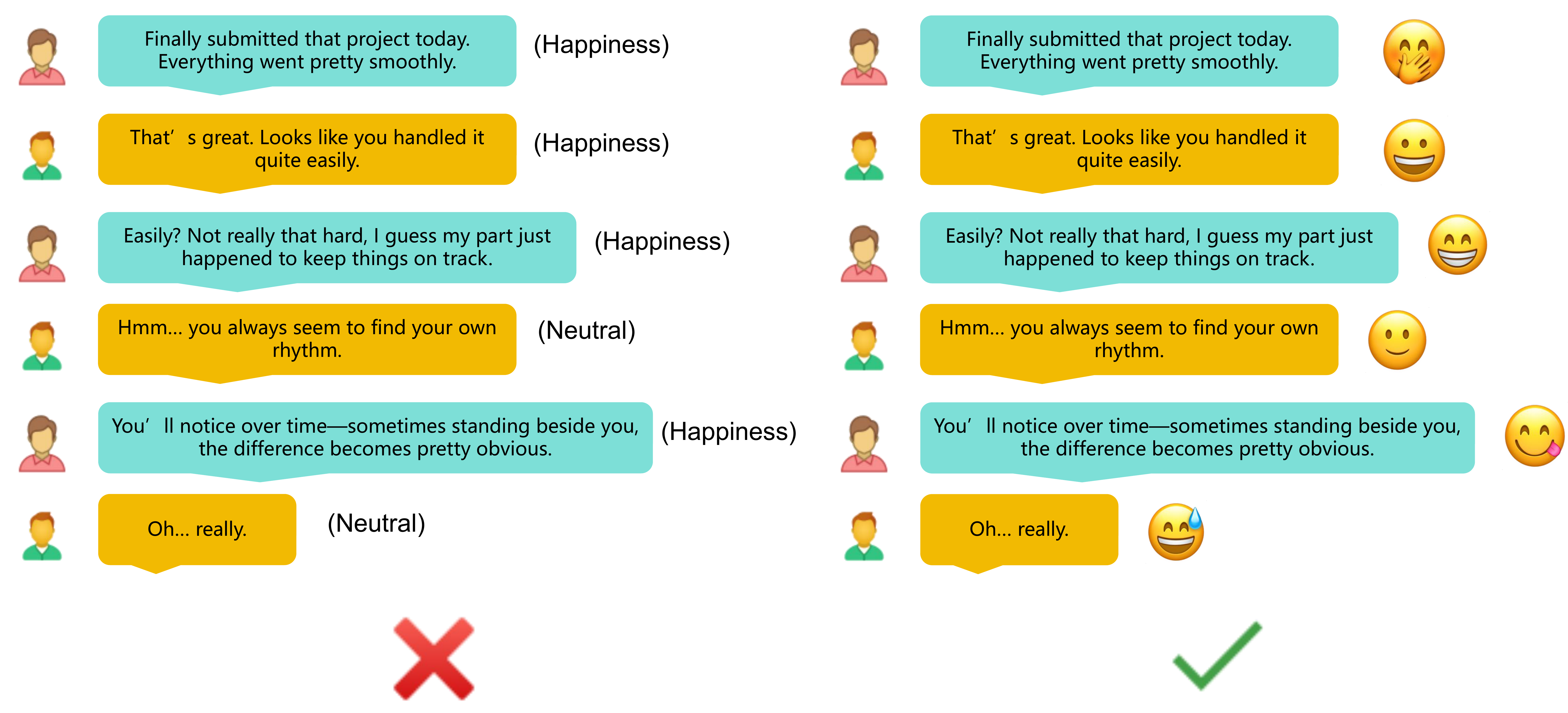}
    \caption{The same coarse emotion label can hide different response-side
    affective-orientation cues. Emoji are used only as weak
    affective--attitudinal contextual signals, not as gold emotion or
    listener-stance labels.}
    \label{fig:emoji-context}
\end{figure}

Figure~\ref{fig:emoji-context} illustrates this distinction. Contexts with the
same coarse emotion label may require different affective uptake, interpersonal
distance, or response strength. Moreover, response-side affective orientation is
often ambiguous: multiple responses may be plausible for the same context, and
annotators may reasonably prefer different orientations. This makes hard,
single-label supervision ill-suited for fine-grained response-orientation
control.

We use emoji as weak supervision for inducing this control
representation. The key point is not that empathetic systems
should generate emoji. Prior work shows that large-scale emoji
prediction can yield transferable representations for sentiment,
emotion, and sarcasm \citep{felbo-etal-2017-using};
textual descriptions can support semantic emoji representations
\citep{eisner-etal-2016-emoji2vec}; and emoji sequences can
express compositional meanings
\citep{yang-etal-2024-elco}.
In our annotation setting, these signals may operationally
correspond to response orientations such as encouragement,
sympathy, celebration, hesitation, teasing, surprise, or concern.
Since emoji meanings are context-dependent and annotators may
disagree, we aggregate multi-annotator emoji votes and confidence
scores into soft distributions, following disagreement-aware
soft-label learning and emotion-distribution estimation
\citep{fornaciari-etal-2021-beyond,
wu-etal-2024-handling}.
We do not treat emoji as output symbols, gold emotion labels,
or gold listener-stance labels. These distributions provide weak
affective--attitudinal evidence for modeling source-side affective
expression and response-side affective orientation.

Based on this idea, we construct \textsc{EmojiDialogue}, an utterance-level
emoji-weakly-supervised extension of EmpatheticDialogues
\citep{rashkin-etal-2019-towards}, where multiple LLM annotators provide emoji
votes and confidence scores for each utterance. We further propose
\textsc{EmoStance}, a controllable generation framework that induces a
name-free latent affective-orientation space, predicts the response-side
orientation from dialogue context and speaker-role transitions, reconstructs it
as a continuous control vector, and injects it into a frozen instruction-tuned
LLM through learned prefix embeddings \citep{li-liang-2021-prefix}. Experiments
show that prototype-based reconstruction is substantially more stable than
direct vector regression. Blind human evaluation further indicates that
\textsc{EmoStance} mainly improves contextual specificity and perceived
responsiveness, while remaining complementary to commonsense-enhanced systems.

Our contributions are threefold. First, we formulate empathetic response
generation as response-side affective-orientation control, using the learned
orientation representation as an operational approximation of listener stance
rather than as a directly annotated stance label. Second, we introduce
\textsc{EmojiDialogue}, a scalable weak-supervision resource that preserves
ambiguity through multi-annotator emoji distributions. Third, we propose
\textsc{EmoStance}, a latent affective-orientation control framework that models
source-side affective expression, predicts context- and role-conditioned
response-side affective orientation, and realizes the predicted orientation
through continuous prefix control of a frozen LLM.
\section{Related Work}

\paragraph{Empathetic and supportive dialogue generation.}
Empathetic dialogue generation is commonly framed as recognizing an interlocutor's affective state and producing an appropriate response, with EmpatheticDialogues serving as a widely used benchmark for emotionally grounded open-domain conversations \citep{rashkin-etal-2019-towards}. 
Early work improves empathetic response generation by incorporating affective signals, such as explicit emotion conditioning, continuous affect representations, and emotion distributions \citep{zhou-etal-2018-emotional,lin-etal-2019-moel,majumder-etal-2020-mime,li-etal-2020-empdg}. 
Another line of work extends empathetic dialogue generation with commonsense cognition or models emotional support conversations through support strategies and user states \citep{liu-etal-2021-towards,sabour-etal-2022-cem,zhao-etal-2023-transesc,zhou-etal-2023-case,li-etal-2024-helpful}. 
More recent studies further introduce discourse-level planning or intent-oriented intermediate variables for supportive response generation \citep{wan-etal-2025-emodynamix,zhang-etal-2025-intentionesc}. 
These studies show the value of explicit intermediate planning, but their planning variables are usually speaker emotions, support strategies, intentions, or external commonsense. 
In contrast, we focus on \emph{listener stance}: the response-side affective and interpersonal orientation that the next utterance should adopt before it is verbalized.

\paragraph{Listener stance and emoji weak supervision.}
Our formulation is related to work on interpersonal stancetaking, which views conversational meaning as a way of positioning the speaker toward the interlocutor, the topic, and the ongoing interaction \citep{kiesling-etal-2018-interactional}. We study a response-side variant of this problem: listener stance describes how the next speaker should take up the previous turn, rather than what private emotion the previous speaker has. Since stance interpretation is subjective, context-dependent, and often underdetermined, our work also follows recent studies arguing that annotator disagreement should be preserved rather than collapsed into a single hard label \citep{fornaciari-etal-2021-beyond,davani-etal-2022-dealing,uma-etal-2021-learning,wu-etal-2024-handling}. Predictive uncertainty has also been used to identify ambiguous
instances in subjective annotation tasks
\citep{alies-etal-2025-measuring}. Emoji provide a useful weak-supervision interface for this purpose because they are compact affective and semantic signals that can express nuanced and sometimes ambiguous interpersonal meanings \citep{eisner-etal-2016-emoji2vec,felbo-etal-2017-using,yang-etal-2024-elco}. Unlike prior emoji-supervised response generation such as MojiTalk \citep{zhou-wang-2018-mojitalk}, we do not predict or output emoji, nor do we use a single emoji as a discrete control code. Instead, we aggregate multi-annotator emoji votes and confidence scores into soft distributions and use them to induce a name-free latent stance space for listener-stance planning.

\paragraph{Continuous control for frozen language models.}
Controllable generation methods steer language models with discrete labels, attribute classifiers, decoding-time discriminators, natural-language instructions, or continuous prompts \citep{pascual-etal-2021-plug-play,yang-klein-2021-fudge,krause-etal-2021-gedi-generative,li-liang-2021-prefix,zhou-etal-2024-evaluating}. EmoStance follows the continuous-control direction, but its control signal is not a manually specified attribute, a binary discriminator target, or a verbal instruction. It is a continuous listener-stance vector induced from emoji weak supervision, predicted from the dialogue context and role transition, and injected into a frozen instruction-tuned LLM through learned prefix embeddings. A more detailed discussion of related work is provided in Appendix~\ref{app:additional_related}.

\section{Weak Affective-Orientation Supervision}
\label{sec:weak-supervision}

We construct \textsc{EmojiDialogue} as an utterance-level weak supervision
layer on top of EmpatheticDialogues~\citep{rashkin-etal-2019-towards}. Since
the original situation-level emotion labels do not specify how a target
response should be affectively and interpersonally oriented, we collect
multi-annotator emoji votes for each utterance and aggregate them into soft
emoji distributions.

We convert adjacent dialogue turns into source--response examples, where the
input contains the situation, dialogue history, and next-speaker marker, and
the target is the next utterance. The resulting \textsc{EmojiDialogue} dataset
comprises 76,489 source--response examples, split into
58,829/9,263/8,397 train/validation/test instances. A human plausibility audit
shows high weak-label plausibility, with 99.69\% valid annotations and 99.77\%
valid or ambiguous-but-acceptable annotations. Full construction, audit,
licensing, privacy, and release details are provided in
Appendices~\ref{app:dataset-details} and~\ref{app:reproducibility}.

\section{Method: Emoji-Supervised Affective-Orientation Control}
\label{sec:method}

EmoStance uses the weak supervision described in
Section~\ref{sec:weak-supervision} to learn an internal control variable for
empathetic response generation. The key idea is to separate two decisions that
are usually entangled in direct generation: how the next response should be
affectively and interpersonally oriented toward the previous speaker, and how
that orientation should be realized in natural language. We call this
intermediate signal a response-side affective-orientation representation.
Emoji annotations are used only during training as weak
affective--attitudinal observations; they are not treated as output symbols,
gold emotion labels, or gold listener-stance labels. At inference time,
EmoStance receives only the dialogue context and next-speaker marker, without
emoji annotations, response-derived orientation vectors, gold listener-stance
labels, or the gold response.

\begin{figure*}[t]
    \centering
    \includegraphics[width=\textwidth]{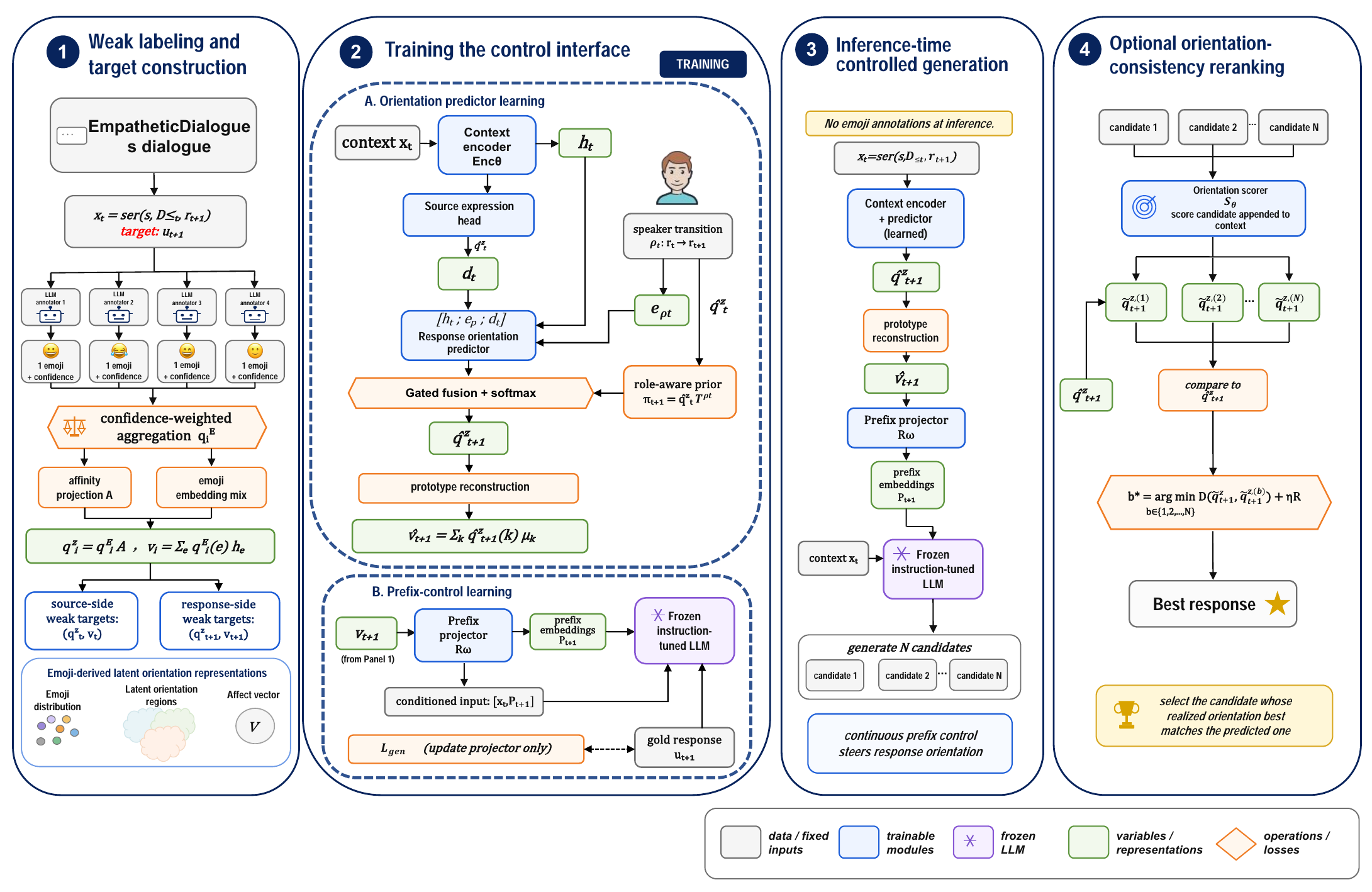}
    \caption{Overview of \textsc{EmoStance}. Emoji weak supervision induces a name-free
latent affective-orientation space for source-side expression modeling,
role-aware response-orientation prediction, prefix-based generation control,
and optional orientation-consistency reranking. Emoji are not treated as gold
emotion or listener-stance labels.}
    \label{fig:method}
\end{figure*}

\subsection{Overview and Problem Setup}
\label{sec:task-formulation}

Let a dialogue be a sequence of role-marked utterances
\[
D=\{(r_1,u_1),\ldots,(r_T,u_T)\},
\]
where \(u_t\) is the utterance at turn \(t\), and \(r_t\) denotes the speaker
role. In dyadic dialogue, \(r_t\in\{A,B\}\), although the formulation also
allows a larger finite set of roles. Each dialogue may additionally include a
situation description \(s\). For each adjacent pair \((u_t,u_{t+1})\), we
construct the serialized input
\[
x_t=\operatorname{ser}(s,D_{\leq t},r_{t+1}),
\]
where \(D_{\leq t}=\{(r_1,u_1),\ldots,(r_t,u_t)\}\), and
\(\operatorname{ser}(\cdot)\) serializes the situation, dialogue history, and
next-speaker marker into a role-marked textual input. The next response is
\(u_{t+1}\).

EmoStance has three stages. First, multi-annotator emoji distributions are
projected into a soft, name-free affective-orientation space, avoiding
predefined orientation names such as happiness, sadness, comfort, or surprise.
Second, a role-aware orientation predictor estimates the response-side
affective-orientation distribution from dialogue context and speaker
transition, conditioned on the source-side affective expression of the latest
observed turn. Third, the predicted distribution is reconstructed through
orientation prototypes into a continuous control vector, which is mapped into
prefix embeddings to steer a frozen instruction-tuned language model. An
optional reranker selects the candidate response whose realized orientation
best matches the predicted orientation.

\subsection{Inducing a Name-Free Affective-Orientation Space}
\label{sec:latent-stance-space}

Affective orientation in dialogue is fine-grained, context-dependent, and often
ambiguous, so we do not represent it as a single hard label. A single utterance
may express several subtle affective or interpersonal cues at once, such as
sympathy, reassurance, cautious encouragement, hesitation, or concern.
EmoStance therefore represents emoji-derived affective--attitudinal evidence
as a soft distribution over latent regions.

For each utterance \(u_t\), annotators select emoji from a fixed candidate
inventory \(\mathcal{E}\) and provide confidence scores. We aggregate these
annotations into a soft emoji distribution
\[
q_t^E \in \Delta^{|\mathcal{E}|}.
\]
This distribution preserves annotator disagreement and confidence variation
rather than collapsing them into a majority label. Such disagreement is not
treated simply as noise: for affective and interpersonal meanings, it may
reflect genuine ambiguity in how an utterance can be read or how a listener
might respond.

Directly using individual emoji as orientation labels would be brittle because
emoji are surface symbols: some are rare, nearly synonymous in a given context,
or polysemous across contexts, and their human-readable names can impose
misleading categories. We therefore induce a name-free affective-orientation
space from relational evidence among emoji, including which emoji appear in
similar textual contexts, are confused or co-selected by annotators, or behave
similarly as weak affective--attitudinal signals.

Concretely, we construct an emoji affinity structure over \(\mathcal{E}\) and
derive a soft membership matrix
\[
A\in[0,1]^{|\mathcal{E}|\times K},
\]
where \(A_{e,k}\) measures the degree to which emoji \(e\) belongs to latent
affective-orientation region \(k\). We write emoji distributions as row vectors
when multiplying by \(A\). The utterance-level latent-region distribution is
\[
q_t^Z = q_t^E A,
\]
where \(q_t^Z\in\Delta^K\). Here, \(Z\) denotes latent emoji-induced
affective-orientation regions, not a gold listener-stance label space. This
projection changes sparse emoji supervision into a smoother distribution over
latent regions. The induced regions act as denoising anchors, retaining
fine-grained information from emoji weak supervision while reducing sensitivity
to idiosyncratic or low-frequency emoji choices.

A distribution over latent regions is useful for orientation prediction, but
generation also benefits from a continuous signal that can express
within-region nuance. Let \(\mathbf{h}_e\) denote the embedding of emoji
\(e\). We define the continuous emoji-derived affective vector for utterance
\(u_t\) as
\[
v_t=\sum_{e\in\mathcal{E}} q_t^E(e)\mathbf{h}_e .
\]
Thus, \(q_t^Z\) provides a structured distribution over latent
affective-orientation regions, while \(v_t\) preserves fine-grained information
from the original emoji distribution.

The same emoji-derived quantities are interpreted according to their role in an
adjacent-turn example. For the latest observed turn \(u_t\),
\((q_t^Z,v_t)\) represents the source-side affective expression available in
the dialogue context. For the target response \(u_{t+1}\),
\((q_{t+1}^Z,v_{t+1})\) represents the response-side affective orientation
observed during training. The central learning problem is to predict the
response-side orientation from the context before generating \(u_{t+1}\).
These representations are induced from weak emoji-based
affective--attitudinal evidence and should not be interpreted as gold
listener-stance labels.

\subsection{Predicting the Response-Side Affective Orientation}
\label{sec:stance-predictor}
\label{sec:stance-vector-reconstruction}

Given the serialized context \(x_t\), \textsc{EmoStance} predicts the
response-side affective orientation that should guide the next response.
Instead of directly regressing a high-dimensional continuous orientation vector
from text, we first predict a distribution over induced latent regions and then
reconstruct a continuous control vector from orientation prototypes. This keeps
prediction tied to the denoised latent affective-orientation space introduced
in Section~\ref{sec:latent-stance-space}.

We encode the context and estimate the source-side affective expression of the
latest observed turn:
\[
\mathbf{h}_t=\operatorname{Enc}_{\theta}(x_t),
\qquad
\hat{q}_t^Z=\operatorname{softmax}(f_{\mathrm{cur}}(\mathbf{h}_t)).
\]
Let \(d_t\) be a compact representation of this predicted source-side
affective expression, and let \(\rho_t=r_t\rightarrow r_{t+1}\) denote the
ordered speaker transition with embedding \(\mathbf{e}_{\rho_t}\). A neural prediction head first produces prior-free, context-based logits for the next-response orientation:
\[
\ell_{t+1}^{0}
=
f_{\mathrm{next}}
\left(
[
\mathbf{h}_t;
\mathbf{e}_{\rho_t};
d_t
]
\right).
\]

The neural predictor captures what the local context suggests, but dyadic
dialogue also exhibits regular affective-uptake patterns: anxious turns often
invite reassurance, celebratory turns invite congratulations, and
self-deprecating turns may invite encouragement or gentle correction. To
incorporate such structure, we estimate a role-aware transition prior from
training data. For each ordered role transition \(\rho\), we estimate a
smoothed transition matrix \(T^\rho\) over latent affective-orientation
regions. Given the predicted source-side expression, the role-conditioned
prior for the next response is
\[
\pi_{t+1}=\hat{q}_t^Z T^{\rho_t}.
\]
The final orientation distribution combines the neural logits and the prior
through a gated interpolation:
\[
\hat{q}_{t+1}^Z
=
\operatorname{softmax}
\left(
\ell_{t+1}^{0}
+
\lambda_{\mathrm{tr}}\gamma_t
\log(\pi_{t+1}+\epsilon)
\right),
\]
where \(\gamma_t\in[0,1]\) controls how strongly the transition prior is used.
The prior is therefore a soft structural bias rather than a replacement for
contextual prediction.

The predicted orientation distribution is then mapped back to a continuous
control vector through orientation prototypes:
\[
\hat{v}_{t+1}
=
\sum_{k=1}^{K}
\hat{q}_{t+1}^Z(k)\mu_k .
\]
This prototype reconstruction preserves continuous control while preventing
the model from chasing idiosyncratic noise in weak emoji-derived vectors.

The orientation predictor is trained with weak supervision derived from the
observed response:
\[
\begin{aligned}
\mathcal{L}_{\mathrm{orient}}
={}&
\operatorname{CE}
\left(
q_{t+1}^Z,
\hat{q}_{t+1}^Z
\right) \\
&+
\lambda_{\mathrm{vec}}
\left\|
\hat{v}_{t+1}-v_{t+1}
\right\|_2^2 \\
&+
\lambda_{\mathrm{cur}}
\mathcal{L}_{\mathrm{cur}} .
\end{aligned}
\]
Here, \(\mathcal{L}_{\mathrm{cur}}\) supervises the source-side expression
estimate \(\hat{q}_t^Z\), which anchors the role-aware transition prior.
Implementation details, including smoothed transition counts, the uncertainty
gate, optional weighted cross-entropy, and the auxiliary prior-free
next-response loss, are given in Appendix~\ref{app:method-details}.

\subsection{Realizing the Predicted Orientation with a Frozen Generator}
\label{sec:generation}

The predicted response-side affective orientation must be realized as natural
language. One possible approach is to verbalize it as a textual prompt, such as
``respond supportively'' or ``sound encouraging.'' We avoid this because the
orientation representation is soft and mixture-like: it may combine several
latent regions with different weights, which is difficult to express as a short
and stable instruction. EmoStance instead uses embedding-level control. In our experiments, the frozen generator is instantiated as
\texttt{mistralai/Mistral-7B-Instruct-v0.3}, from the Mistral~7B
model family \citep{jiang-etal-2023-mistral}. Exact model checkpoints,
parameter counts, licenses, and compute details are reported in
Appendix~\ref{app:reproducibility}.

A lightweight prefix projector maps an orientation vector into \(m\) continuous
prefix embeddings:
\[
P_{t+1}
=
R_{\omega}(v_{t+1})
\in
\mathbb{R}^{m\times d_{\Omega}},
\]
where \(d_{\Omega}\) is the embedding dimension of the frozen generator. The
prefix embeddings are prepended to the serialized dialogue input, allowing the
orientation signal to guide generation without appearing as explicit text.

During projector training, the observed response \(u_{t+1}\) is available, so
we use its weak response-side orientation vector \(v_{t+1}\). The generator
parameters \(\Omega\) remain frozen, and only the prefix projector parameters
\(\omega\) are updated:
\[
\mathcal{L}_{\mathrm{gen}}
=
-
\sum_{j=1}^{|u_{t+1}|}
\log
p_{\Omega}
\left(
u_{t+1,j}
\mid
P_{t+1},
x_t,
u_{t+1,<j}
\right).
\]
At inference time, \(v_{t+1}\) is replaced by the predicted vector
\(\hat{v}_{t+1}\). The frozen language model then generates from
\[
p_{\Omega}
\left(
\cdot
\mid
R_{\omega}(\hat{v}_{t+1}),
x_t
\right).
\]
This design separates orientation prediction from surface realization: the
orientation predictor decides how the next response should be positioned,
while the frozen generator realizes that control signal in natural language.

\subsection{Orientation-Consistency Reranking}
\label{sec:stance-reranking}

Prefix control does not guarantee that every sampled response realizes the
intended orientation. The frozen generator may still produce several fluent but
differently positioned continuations. EmoStance therefore uses
orientation-consistency reranking as an optional decoding-time check.

Given the predicted orientation \(\hat{q}_{t+1}^Z\) and control vector
\(\hat{v}_{t+1}\), the generator samples \(B\) candidate responses:
\[
\left\{
\widetilde{u}_{t+1}^{(1)},
\ldots,
\widetilde{u}_{t+1}^{(B)}
\right\}.
\]
For each candidate, we append it to the dialogue context and use the
orientation scorer to estimate the orientation realized by that candidate:
\[
\widetilde{q}_{t+1}^{Z,(b)}
=
S_{\theta}
\left(
x_t,
\widetilde{u}_{t+1}^{(b)}
\right).
\]
The selected response minimizes divergence from the intended orientation, with
an optional length regularizer \(\mathcal{R}\):
\[
b^*
=
\arg\min_{1\leq b\leq B}
\left[
D
\left(
\hat{q}_{t+1}^Z,
\widetilde{q}_{t+1}^{Z,(b)}
\right)
+
\eta
\mathcal{R}
\left(
\widetilde{u}_{t+1}^{(b)}
\right)
\right],
\]
\[
\qquad
\hat{u}_{t+1}
=
\widetilde{u}_{t+1}^{(b^*)}.
\]
By default, \(D\) is cross-entropy between the intended and realized
orientation distributions. Reranking does not introduce additional labels or
external knowledge; it only selects the candidate whose realized orientation
best matches the predicted orientation.

\subsection{Training and Inference}
\label{sec:training-inference}

Training and inference differ only in the availability of weak
affective-orientation observations. During training, emoji annotations of the
observed source--response pair are aggregated into soft emoji distributions,
projected into the name-free affective-orientation space, and used to train the
orientation predictor and prefix projector with
\(\mathcal{L}_{\mathrm{orient}}\) and \(\mathcal{L}_{\mathrm{gen}}\), while the
generator remains frozen. At inference time, no emoji annotations,
response-derived orientation vectors, gold listener-stance labels, or gold
responses are available. EmoStance predicts the response-side affective
orientation from \(x_t\), reconstructs the corresponding control vector, maps
it into prefix embeddings, and generates with the frozen language model;
optional reranking selects the candidate most consistent with the predicted
orientation. Thus, emoji are used only as training-time weak observations for
learning an internal affective-orientation control interface, not as test-time
inputs, gold emotion labels, gold listener-stance labels, or desired outputs.
% This protocol is central to the interpretation of the method. Emoji are not
% appended to the test-time input, are not treated as gold emotion labels, and
% are not the desired output. They are used only as weak training observations
% for learning a soft stance space and an internal continuous control interface
% for empathetic response generation.
\section{Experiments}
\label{sec:experiments}

We evaluate \textsc{EmoStance} along three questions: whether response-side
affective-orientation control improves empathetic response generation, which
aspects of response quality it affects, and whether the proposed components are
necessary. Blind pairwise human preference is treated as the primary evidence
for generation quality, while automatic metrics and internal
orientation-control diagnostics are used as supporting analyses rather than
substitutes for human judgment.

\subsection{Experimental Setup}
\label{sec:exp-setup}

We evaluate \textsc{EmoStance} in two settings: component analyses on the
\textsc{EmojiDialogue} adjacent-turn split introduced in
Section~\ref{sec:weak-supervision}, and system comparisons on the full
EmpatheticDialogues (ED) test set. In every deployable setting, the input is
restricted to the situation description, dialogue history, and speaker-role
markers. Emoji annotations, latent-region targets, response-derived orientation
vectors, and reference responses are unavailable at inference time. Dataset
statistics and decoding settings are provided in
Appendix~\ref{app:data-inference}.

We compare against seven baselines: an
instruction-tuned LLM without affective control,
a prompt-level control variant, supervised fine-tuning
without the latent orientation module, an EmPO-DPO
preference-optimization baseline
\citep{sotolar2024empo}, two ED-compatible
task-specific systems---CASE
\citep{zhou-etal-2023-case} and APTNESS
\citep{hu-etal-2024-aptness}---and Sibyl, a
future-aware commonsense-enhanced system
\citep{wang-etal-2025-sibyl}. All baseline outputs
are produced by our own ED-compatible reproductions under the shared
\texttt{mistralai/Mistral-7B-Instruct-v0.3} backbone, aligned input format,
test contexts, and decoding setup. These comparisons therefore evaluate
controlled same-backbone variants and should not be interpreted as exact
replications or upper bounds of the original released systems.
Appendix~\ref{app:baseline-details} provides the full baseline configurations.

Our primary evaluation is blind pairwise human preference. Each item presents a
dialogue context, one evaluation question, and anonymized responses from
\textsc{EmoStance} and one baseline. The five dimensions are emotion
appropriateness, felt responded, context specificity, naturalness, and
AI-like/problematic phrasing. The first four are positive dimensions, whereas
AI-like/problematic phrasing is reverse-scored.

We conducted two evaluation batches using the same instructions, blinding, and
scoring protocol. The second batch recruited 10 new annotators and
independently sampled new evaluation instances rather than reannotating the
original items. The combined evaluation contains 20 annotators and 800
judgments. Ties and ``neither/both bad'' are retained as neutral outcomes and
excluded from decisive win rates. We report 95\% Wilson confidence intervals
and two-sided exact sign tests. The main table additionally reports
Holm-adjusted \(p\)-values across the seven per-baseline comparisons.
Appendix~\ref{app:human-protocol} provides the complete protocol.

\subsection{Human Evaluation}
\label{sec:main-human-eval}

Table~\ref{tab:human-main-baseline} reports the expanded blind pairwise
evaluation. Across 800 judgments, \textsc{EmoStance} receives 395 wins, 71
ties, 94 neither/both-bad judgments, and 240 losses. Excluding neutral
outcomes, the overall decisive win rate is 62.2\%, with a 95\% Wilson
confidence interval of [58.4, 65.9] and a two-sided exact sign-test value of
\(p<.001\).

\begin{table*}[t]
\centering
\small
\setlength{\tabcolsep}{4pt}
\begin{tabular}{lrrrrrrrr}
\toprule
Baseline
& Win & Tie & None & Lose
& Win\% & 95\% CI & \(p\) & \(p_{\mathrm{Holm}}\) \\
\midrule
LLM-only
& 49 & 17 & 13 & 41
& 54.4 & [44.2, 64.3] & 0.461 & 0.754 \\

LLM-prompt
& 52 & 12 & 16 & 40
& 56.5 & [46.3, 66.2] & 0.251 & 0.754 \\

LLM-SFT
& 62 & 10 & 11 & 37
& 62.6 & [52.8, 71.5] & 0.015 & 0.077 \\

EmPO-DPO
& 54 & 13 & 14 & 39
& 58.1 & [47.9, 67.6] & 0.146 & 0.585 \\

CASE
& 80 & 3 & 15 & 22
& 78.4 & [69.5, 85.3] & \(<.001\) & \(<.001\) \\

APTNESS
& 72 & 8 & 14 & 26
& 73.5 & [64.0, 81.2] & \(<.001\) & \(<.001\) \\

Sibyl
& 26 & 8 & 11 & 35
& 42.6 & [31.0, 55.1] & 0.306 & 0.754 \\
\midrule
Overall
& 395 & 71 & 94 & 240
& 62.2 & [58.4, 65.9] & \(<.001\) & -- \\
\bottomrule
\end{tabular}
\caption{
Expanded blind pairwise human evaluation. Wins and losses are counted from
\textsc{EmoStance}'s perspective; ties and neither/both-bad outcomes are
excluded from Win\%. The \(p\) column reports uncorrected two-sided exact sign
tests over decisive judgments. \(p_{\mathrm{Holm}}\) applies Holm correction
across the seven per-baseline comparisons.
}
\label{tab:human-main-baseline}
\end{table*}

The aggregate result is qualified by the per-baseline comparisons. After Holm
correction, \textsc{EmoStance} is significantly preferred over the controlled
CASE and APTNESS variants. The numerical margin over LLM-SFT is positive but
does not remain significant after correction, and the comparisons with
LLM-only, LLM-prompt, EmPO-DPO, and Sibyl are statistically inconclusive. In
particular, Sibyl receives more decisive preferences numerically, but its
confidence interval includes parity. We therefore do not claim uniform
dominance over strong or knowledge-enhanced systems. The evidence instead
supports response-side affective orientation as a useful control signal that
improves aggregate preference and is potentially complementary to preference
optimization and future-aware commonsense modeling.

Dimension-level results identify where the aggregate gain arises.
\textsc{EmoStance} achieves decisive win rates of 75.9\% for context
specificity and 73.5\% for felt responded. The estimates for emotion
appropriateness (56.0\%) and naturalness (57.7\%) are more modest, and their
confidence intervals include parity. AI-like/problematic phrasing is also
statistically inconclusive at 45.1\%. Thus, the supported improvement is
concentrated in contextual uptake and perceived responsiveness rather than
broad surface-form enhancement. Full counts and confidence intervals are
reported in Appendix~\ref{app:human-dimension-results}.

To assess the weak supervision separately from response preference, we also
conduct a human--LLM distributional audit on 120 test utterances. Exact emoji
choices show non-trivial divergence between human and LLM annotators, whereas
projection through the fixed learned emoji-to-region matrix reduces mean JSD
from 0.442 to 0.206 and increases distributional overlap from 0.449 to 0.670.
We interpret this as sample-specific evidence that the induced orientation
regions are less sensitive to exact-symbol variation, not as evidence that LLM
annotations are equivalent to human annotations or culturally universal.
Appendix~\ref{app:human-llm-audit} provides the complete audit.

\subsection{Automatic Evaluation}
\label{sec:automatic-eval}

Appendix~D.6 reports reference-based metrics---
\mbox{BERTScore-F1}, \mbox{ROUGE-L}, \mbox{BLEU-2},
and METEOR
\citep{Zhang20:bertscore,
lin-2004-rouge,
papineni-etal-2002-bleu,
banerjee-lavie-2005-meteor}---
together with \mbox{Distinct-1/2}
\citep{li-etal-2016-diversity},
\mbox{Self-BLEU}
\citep{zhu-etal-2018-texygen},
and a rule-based Generic diagnostic. On the aligned ED test set,
\textsc{EmoStance} obtains the highest BERTScore-F1, ROUGE-L, and BLEU-2 among
the controlled same-backbone systems. METEOR and diversity-related diagnostics
are mixed: \textsc{EmoStance} is not the most lexically diverse system and
does not obtain the lowest generic-response rate. We therefore interpret these
metrics narrowly as reference-alignment and surface-form diagnostics. They do
not by themselves establish superior empathy, naturalness, diversity, or
reduced template-like phrasing.

\subsection{Component Analysis}
\label{sec:component-analysis}

We examine response-orientation prediction, continuous orientation-vector
construction, and generation-time control; full tables and secondary
diagnostics are provided in Appendices~\ref{app:auto-ablation}
and~\ref{app:supp-ablation}.

\begin{table*}[t]
\centering
\small
\setlength{\tabcolsep}{5pt}
\begin{tabular}{lrrrrrrr}
\toprule
Comparison
& Win & Tie & Neither & Lose
& Win Rate & 95\% CI & \(p\) \\
\midrule
vs.\ w/o rerank
& 156 & 53 & 18 & 73
& 68.1\% & [61.8, 73.8] & \(<.001\) \\

vs.\ w/o role-aware
& 138 & 66 & 18 & 78
& 63.9\% & [57.3, 70.0] & \(<.001\) \\

vs.\ zero control
& 222 & 20 & 21 & 37
& 85.7\% & [80.9, 89.5] & \(<.001\) \\
\midrule
Overall
& 516 & 139 & 57 & 188
& 73.3\% & [69.9, 76.4] & \(<.001\) \\
\bottomrule
\end{tabular}
\caption{
Expanded focused human ablation at the judgment level. Wins and losses are
counted from the final \textsc{EmoStance} system's perspective; ties and
neither/both-bad outcomes are excluded from the decisive win rate.
}
\label{tab:human-ablation}
\end{table*}

First, soft distributional supervision is more effective than argmax targets,
improving response-orientation prediction from 1.4450 to 1.3792 in CE and from
0.3067 to 0.3260 in macro-F1. Second, prototype reconstruction is substantially
more stable than direct 256-dimensional regression, increasing target-vector
cosine similarity from 0.3220 to 0.9236 and reducing MSE from 0.001058 to
0.000022. This result supports the predictability of the prototype-structured
control representation, but it should not be interpreted as evidence that the
prototype mixture preserves all information in the dense response-derived
target.

Third, generation-control diagnostics show that predicted orientation controls
are more meaningful than shuffled controls and that reranking improves
realization of the supplied orientation. The remaining gap to
reference-conditioned upper-reference settings reflects both prediction error
and task underdetermination: the observed context does not uniquely determine
how a listener must respond, while the upper-reference conditions directly
observe the orientation realized in the single ED reference response. We
therefore treat the reference response as one plausible human continuation
rather than the unique correct orientation.

Table~\ref{tab:human-ablation} reports the expanded human ablation. The final
system is preferred over all three deployable variants, with decisive win rates
of 68.1\% over no reranking, 63.9\% over the variant without role-aware
response-orientation prediction, and 85.7\% over zero control. These results
support the contributions of the orientation signal, role-aware prediction, and
orientation-consistency selection.

Reranking has a measurable efficiency cost. On a single RTX 4090 under matched
decoding settings, the \(B=1\) no-reranking configuration requires 331.7 ms per
example on average and processes 3.015 examples/s, whereas \(B=4\) reranking
requires 1,333.4 ms and processes 0.750 examples/s, corresponding to a
\(4.02\times\) cost increase. Candidate generation accounts for 99.48\% of the
\(B=4\) runtime, while orientation scoring and final selection together account
for only 0.52\%. We therefore present \(B=1\) as the efficiency-oriented
deployment mode and \(B=4\) as the quality-oriented mode.
Appendix~\ref{app:inference-efficiency} provides the complete latency and
profiling results.
\section{Conclusion}
\label{sec:conclusion}

We introduced \textsc{EmoStance}, a weakly supervised and role-aware framework
for listener-stance control in empathetic response generation. The method uses
multi-annotator emoji distributions as soft training-time signals, predicts a
distribution over plausible next-response stances, and reconstructs a
prototype-structured continuous control vector that steers a frozen
instruction-tuned language model through prefix embeddings. Blind pairwise
evaluation with 20 annotators and 800 judgments yields a 62.2\% aggregate
decisive win rate, with the clearest gains in context specificity and felt
responded. These findings support listener stance as a useful intermediate
variable for improving contextual uptake and perceived responsiveness, but do
not indicate uniform superiority across all quality dimensions or strong
baselines. Future work should improve uncertainty-aware stance prediction,
develop richer and more efficient control mechanisms, and evaluate
generalization across languages, cultures, datasets, and interaction settings.
\section*{Limitations}
\paragraph{Weak supervision and construct validity.}
The response-side affective-orientation targets are derived from
LLM-provided emoji annotations rather than direct human labels of emotion,
empathy, mental state, or listener stance. Human audits support their
contextual plausibility, but do not establish gold-standard, exhaustive, or
culturally universal supervision. Emoji meanings vary across communities,
platforms, age groups, and conversational norms, and the name-free design
cannot remove biases inherited from the source corpus or annotator models.
The induced space should therefore be interpreted as a corpus-dependent
control representation that approximates aspects of listener stance, not as
an independently validated taxonomy.

\paragraph{Scope and evaluation.}
Experiments are limited to short English dyadic conversations from
EmpatheticDialogues and do not cover long-horizon support, multi-party
interaction, persistent memory, or open-domain assistants with stronger
factual, safety, and tool-use requirements. The appropriate response
orientation may also be underdetermined: several orientations can be
reasonable for the same context, so the gap to reference-conditioned
upper-reference settings reflects both prediction error and task ambiguity.
Moreover, several automatic consistency metrics reuse the induced
orientation space or a related scorer and should be treated as internal
control-realization diagnostics rather than independent evidence of empathy.
The human studies evaluate static response pairs rather than live,
longitudinal interactions.

\paragraph{Efficiency and capability boundaries.}
Multi-candidate orientation-consistency reranking improves control at higher
inference cost, while single-generation decoding provides a lower-cost
alternative. \textsc{EmoStance} targets affective and interpersonal
orientation rather than commonsense reasoning, factual grounding, safety, or
surface-form quality; its contribution is therefore complementary to
knowledge-augmented and other optimization methods.
\section*{Ethical Considerations}

\paragraph{Interpretation and annotation bias.}
Emoji annotations and latent orientations are weak conversational signals,
not ground-truth emotions, psychological states, personality traits,
clinical indicators, or evidence of a user's internal feelings. Generated
responses reflect a selected communicative orientation rather than a
diagnosis. Although soft distributions preserve disagreement, LLM annotators
and emoji conventions may still encode biases involving dialect,
indirectness, politeness, humor, cultural norms, disability-related
communication styles, or non-standard phrasing. The resulting resources
should not be used for user profiling, mental-state detection, clinical
decision making, or authoritative affective judgment.

\paragraph{Deployment risks.}
Affective-orientation control may make systems appear more emotionally
attuned, but it could also be used for persuasion, dependency induction, or
emotional manipulation. Systems should not exploit distress, covertly steer
decisions, simulate human care relationships, or override safety policies.
\textsc{EmoStance} is not designed for diagnosis, crisis counseling, medical
or legal advice, or professional emotional care; safety-critical cases
require appropriate refusal, escalation, crisis-resource referral, and human
oversight.

\paragraph{Privacy, transparency, and release.}
Benchmark dialogue may contain sensitive personal experiences. Data
collection, auditing, and release should respect applicable consent,
compensation, licensing, and data-protection requirements, document the
weak-supervision procedure and non-clinical scope, and avoid exposing
identifying information. Users should also be informed when they are
interacting with an AI system and when responses may be guided by inferred
affective or interpersonal orientations.
\bibliography{custom}

@inproceedings{zhang-etal-2025-decoupledesc,
  title={{D}ecoupled{ESC}: Enhancing Emotional Support Generation via Strategy-Response Decoupled Preference Optimization},
    author = "Zhang, Chao  and
      Shi, Xin  and
      Zhang, Xueqiao  and
      Zhu, Yifan  and
      Yang, Yi  and
      Luo, Yawei",
    editor = "Christodoulopoulos, Christos  and
      Chakraborty, Tanmoy  and
      Rose, Carolyn  and
      Peng, Violet",
  booktitle={Findings of the Association for Computational Linguistics: EMNLP 2025},
    month = nov,
    year = "2025",
    address = "Suzhou, China",
    publisher = "Association for Computational Linguistics",
    url = "https://aclanthology.org/2025.findings-emnlp.1209/",
    doi = "10.18653/v1/2025.findings-emnlp.1209",
    pages = "22189--22215",
    ISBN = "979-8-89176-335-7",
}

@inproceedings{zhou-etal-2024-evaluating,
  title={Evaluating the Smooth Control of Attribute Intensity in Text Generation with {LLM}s},
  author={Zhou, Shang and Yao, Feng and Dong, Chengyu and Wang, Zihan and Shang, Jingbo},
  booktitle={Findings of the Association for Computational Linguistics: ACL 2024},
  pages={4348--4362},
  year={2024}
}

@inproceedings{wan-etal-2025-emodynamix,
  title={{EmoDynamiX}: Emotional Support Dialogue Strategy Prediction by Modelling {MiXed} Emotions and Discourse Dynamics},
  author={Wan, Chenwei and Labeau, Matthieu and Clavel, Chlo{\'e}},
  booktitle={Proceedings of the 2025 Conference of the Nations of the Americas Chapter of the Association for Computational Linguistics: Human Language Technologies (Volume 1: Long Papers)},
  pages={1678--1695},
  year={2025}
}

@inproceedings{colombo-etal-2019-affect-driven,
  title={Affect-Driven Dialog Generation},
  author={Colombo, Pierre and Witon, Wojciech and Modi, Ashutosh and Kennedy, James and Kapadia, Mubbasir},
  booktitle={Proceedings of the 2019 Conference of the North American Chapter of the Association for Computational Linguistics: Human Language Technologies, Volume 1 (Long and Short Papers)},
  pages={3734--3743},
  year={2019}
}

@inproceedings{mohammad-2018-obtaining,
  title={Obtaining Reliable Human Ratings of Valence, Arousal, and Dominance for 20,000 {English} Words},
  author={Mohammad, Saif},
  booktitle={Proceedings of the 56th Annual Meeting of the Association for Computational Linguistics (Volume 1: Long Papers)},
  pages={174--184},
  year={2018}
}

@inproceedings{wu-etal-2024-handling,
  title={Handling Ambiguity in Emotion: From Out-of-Domain Detection to Distribution Estimation},
  author={Wu, Wen and Li, Bo and Zhang, Chao and Chiu, Chung-Cheng and Li, Qiujia and Bai, Junwen and Sainath, Tara and Woodland, Phil},
  booktitle={Proceedings of the 62nd Annual Meeting of the Association for Computational Linguistics (Volume 1: Long Papers)},
  pages={2078--2093},
  year={2024}
}

@inproceedings{fornaciari-etal-2021-beyond,
  title={Beyond Black \& White: Leveraging Annotator Disagreement via Soft-Label Multi-Task Learning},
  author={Fornaciari, Tommaso and Uma, Alexandra and Paun, Silviu and Plank, Barbara and Hovy, Dirk and Poesio, Massimo},
  booktitle={Proceedings of the 2021 Conference of the North American Chapter of the Association for Computational Linguistics: Human Language Technologies},
  pages={2591--2597},
  year={2021}
}

@inproceedings{felbo-etal-2017-using,
  title={Using Millions of Emoji Occurrences to Learn Any-Domain Representations for Detecting Sentiment, Emotion and Sarcasm},
  author={Felbo, Bjarke and Mislove, Alan and S{\o}gaard, Anders and Rahwan, Iyad and Lehmann, Sune},
  booktitle={Proceedings of the 2017 Conference on Empirical Methods in Natural Language Processing},
  pages={1615--1625},
  year={2017}
}

@inproceedings{zhou-wang-2018-mojitalk,
  title={{MojiTalk}: Generating Emotional Responses at Scale},
  author={Zhou, Xianda and Wang, William Yang},
  booktitle={Proceedings of the 56th Annual Meeting of the Association for Computational Linguistics (Volume 1: Long Papers)},
  pages={1128--1137},
  year={2018}
}

@inproceedings{rashkin-etal-2019-towards,
  title={Towards Empathetic Open-Domain Conversation Models: A New Benchmark and Dataset},
  author={Rashkin, Hannah and Smith, Eric Michael and Li, Margaret and Boureau, Y-Lan},
  booktitle={Proceedings of the 57th Annual Meeting of the Association for Computational Linguistics},
  pages={5370--5381},
  year={2019}
}

@inproceedings{zhao-etal-2023-transesc,
  title={{TransESC}: Smoothing Emotional Support Conversation via Turn-Level State Transition},
  author={Zhao, Weixiang and Zhao, Yanyan and Wang, Shilong and Qin, Bing},
  booktitle={Findings of the Association for Computational Linguistics: ACL 2023},
  pages={6725--6739},
  year={2023}
}

@inproceedings{deng-etal-2023-knowledge,
  title={Knowledge-Enhanced Mixed-Initiative Dialogue System for Emotional Support Conversations},
  author={Deng, Yang and Zhang, Wenxuan and Yuan, Yifei and Lam, Wai},
  booktitle={Proceedings of the 61st Annual Meeting of the Association for Computational Linguistics (Volume 1: Long Papers)},
  pages={4079--4095},
  year={2023}
}

@inproceedings{li-etal-2024-helpful,
  title={Be Helpful but Don{'}t Talk too Much - Enhancing Helpfulness in Conversations through Relevance in Multi-Turn Emotional Support},
  author={Li, Junlin and Peng, Bo and Hsu, Yu-Yin and Huang, Chu-Ren},
  booktitle={Proceedings of the 2024 Conference on Empirical Methods in Natural Language Processing},
  pages={1976--1988},
  year={2024},
  doi={10.18653/v1/2024.emnlp-main.118},
  url={https://aclanthology.org/2024.emnlp-main.118/},
  publisher={Association for Computational Linguistics},
  address={Miami, Florida, USA}
}

@inproceedings{zhang-etal-2025-intentionesc,
  title={{IntentionESC}: An Intention-Centered Framework for Enhancing Emotional Support in Dialogue Systems},
  author={Zhang, Xinjie and Wang, Wenxuan and Jin, Qin},
  booktitle={Findings of the Association for Computational Linguistics: ACL 2025},
  pages={26494--26516},
  year={2025}
}

@article{rodriguez-barroso-etal-2024-federated,
  title={Federated Learning for Exploiting Annotators' Disagreements in Natural Language Processing},
  author={Rodr{\'i}guez-Barroso, Nuria and C{\'a}mara, Eugenio Mart{\'i}nez and Collados, Jose Camacho and Luz{\'o}n, M. Victoria and Herrera, Francisco},
  journal={Transactions of the Association for Computational Linguistics},
  volume={12},
  pages={630--648},
  year={2024},
  doi={10.1162/tacl_a_00664},
  url={https://aclanthology.org/2024.tacl-1.35/}
}

@inproceedings{alies-etal-2025-measuring,
  title={Measuring Label Ambiguity in Subjective Tasks using Predictive Uncertainty Estimation},
  author={Alies, Richard and Merdjanovska, Elena and Akbik, Alan},
  booktitle={Proceedings of the 19th Linguistic Annotation Workshop (LAW-XIX-2025)},
  pages={21--34},
  year={2025}
}

@inproceedings{eisner-etal-2016-emoji2vec,
  title={emoji2vec: Learning Emoji Representations from their Description},
  author={Eisner, Ben and Rockt{\"a}schel, Tim and Augenstein, Isabelle and Bo{\v{s}}njak, Matko and Riedel, Sebastian},
  booktitle={Proceedings of the Fourth International Workshop on Natural Language Processing for Social Media},
  pages={48--54},
  year={2016}
}

@inproceedings{yang-etal-2024-elco,
  title={The {ELCo} Dataset: Bridging Emoji and Lexical Composition},
  author={Yang, Zi Yun and Zhang, Ziqing and Miao, Yisong},
  booktitle={Proceedings of the 2024 Joint International Conference on Computational Linguistics, Language Resources and Evaluation (LREC-COLING 2024)},
  pages={15899--15909},
  year={2024}
}

@inproceedings{wang-etal-2025-sibyl,
  title={Sibyl: Empowering Empathetic Dialogue Generation in Large Language Models via Sensible and Visionary Commonsense Inference},
  author={Wang, Lanrui and Li, Jiangnan and Yang, Chenxu and Lin, Zheng and Tang, Hongyin and Liu, Huan and Cao, Yanan and Wang, Jingang and Wang, Weiping},
  booktitle={Proceedings of the 31st International Conference on Computational Linguistics},
  pages={123--140},
  year={2025}
}

@inproceedings{zhou-etal-2018-emotional,
  title={Emotional Chatting Machine: Emotional Conversation Generation with Internal and External Memory},
  author={Zhou, Hao and Huang, Minlie and Zhang, Tianyang and Zhu, Xiaoyan and Liu, Bing},
  booktitle={Proceedings of the AAAI Conference on Artificial Intelligence},
  volume={32},
  number={1},
  pages={730--738},
  year={2018},
  doi={10.1609/aaai.v32i1.11325}
}

@inproceedings{lin-etal-2019-moel,
  title={{MoEL}: Mixture of Empathetic Listeners},
  author={Lin, Zhaojiang and Madotto, Andrea and Shin, Jamin and Xu, Peng and Fung, Pascale},
  booktitle={Proceedings of the 2019 Conference on Empirical Methods in Natural Language Processing and the 9th International Joint Conference on Natural Language Processing (EMNLP-IJCNLP)},
  pages={121--132},
  year={2019}
}

@inproceedings{majumder-etal-2020-mime,
  title={{MIME}: {MIM}icking Emotions for Empathetic Response Generation},
  author={Majumder, Navonil and Hong, Pengfei and Peng, Shanshan and Lu, Jiankun and Ghosal, Deepanway and Gelbukh, Alexander and Mihalcea, Rada and Poria, Soujanya},
  booktitle={Proceedings of the 2020 Conference on Empirical Methods in Natural Language Processing (EMNLP)},
  pages={8968--8979},
  year={2020}
}

@inproceedings{li-etal-2020-empdg,
  title={{EmpDG}: Multi-Resolution Interactive Empathetic Dialogue Generation},
  author={Li, Qintong and Chen, Hongshen and Ren, Zhaochun and Ren, Pengjie and Tu, Zhaopeng and Chen, Zhumin},
  booktitle={Proceedings of the 28th International Conference on Computational Linguistics},
  pages={4454--4466},
  year={2020}
}

@inproceedings{liu-etal-2021-towards,
  title={Towards Emotional Support Dialog Systems},
  author={Liu, Siyang and Zheng, Chujie and Demasi, Orianna and Sabour, Sahand and Li, Yu and Yu, Zhou and Jiang, Yong and Huang, Minlie},
  booktitle={Proceedings of the 59th Annual Meeting of the Association for Computational Linguistics and the 11th International Joint Conference on Natural Language Processing (Volume 1: Long Papers)},
  pages={3469--3483},
  year={2021}
}

@inproceedings{tu-etal-2022-misc,
  title={{MISC}: A Mixed Strategy-Aware Model Integrating {COMET} for Emotional Support Conversation},
  author={Tu, Quan and Li, Yanran and Cui, Jianwei and Wang, Bin and Wen, Ji-Rong and Yan, Rui},
  booktitle={Proceedings of the 60th Annual Meeting of the Association for Computational Linguistics (Volume 1: Long Papers)},
  pages={308--319},
  year={2022}
}

@article{sabour-etal-2022-cem,
  title={{CEM}: Commonsense-Aware Empathetic Response Generation},
  author  = {Sabour, Sahand and Zheng, Chujie and Huang, Minlie},
  journal = {Proceedings of the AAAI Conference on Artificial Intelligence},
  volume  = {36},
  number  = {10},
  pages   = {11229--11237},
  year    = {2022},
  doi     = {10.1609/aaai.v36i10.21373},
  url     = {https://ojs.aaai.org/index.php/AAAI/article/view/21373}
}

@inproceedings{zhou-etal-2023-case,
  title={{CASE}: Aligning Coarse-to-Fine Cognition and Affection for Empathetic Response Generation},
  author={Zhou, Jinfeng and Zheng, Chujie and Wang, Bo and Zhang, Zheng and Huang, Minlie},
  booktitle={Proceedings of the 61st Annual Meeting of the Association for Computational Linguistics (Volume 1: Long Papers)},
  pages={8223--8237},
  year={2023}
}

@article{kiesling-etal-2018-interactional,
  title={Interactional Stancetaking in Online Forums},
  author={Kiesling, Scott F. and Pavalanathan, Umashanthi and Fitzpatrick, Jim and Han, Xiaochuang and Eisenstein, Jacob},
  journal={Computational Linguistics},
  volume={44},
  number={4},
  pages={683--718},
  year={2018}
}

@article{davani-etal-2022-dealing,
  title   = {Dealing with Disagreements: Looking Beyond the Majority Vote in Subjective Annotations},
  author  = {Mostafazadeh Davani, Aida and
             D{\'i}az, Mark and
             Prabhakaran, Vinodkumar},
  journal = {Transactions of the Association for Computational Linguistics},
  volume  = {10},
  pages   = {92--110},
  year    = {2022},
  doi     = {10.1162/tacl_a_00449}
}

@article{uma-etal-2021-learning,
  title={Learning from Disagreement: A Survey},
  author={Uma, Alexandra N. and Fornaciari, Tommaso and Hovy, Dirk and Paun, Silviu and Plank, Barbara and Poesio, Massimo},
  journal={Journal of Artificial Intelligence Research},
  volume={72},
  pages={1385--1470},
  year={2021}
}

@article{gilardi2023chatgpt,
  title={{ChatGPT} Outperforms Crowd Workers for Text-Annotation Tasks},
  author={Gilardi, Fabrizio and Alizadeh, Meysam and Kubli, Ma{\"e}l},
  journal={Proceedings of the National Academy of Sciences},
  volume={120},
  number={30},
  pages={e2305016120},
  year={2023},
  doi={10.1073/pnas.2305016120},
  url={https://doi.org/10.1073/pnas.2305016120}
}

@inproceedings{tan-etal-2024-large,
  title={Large Language Models for Data Annotation and Synthesis: A Survey},
  author={Tan, Zhen and Li, Dawei and Wang, Song and Beigi, Alimohammad and Jiang, Bohan and Bhattacharjee, Amrita and Karami, Mansooreh and Li, Jundong and Cheng, Lu and Liu, Huan},
  booktitle={Proceedings of the 2024 Conference on Empirical Methods in Natural Language Processing},
  pages={930--957},
  year={2024}
}

@inproceedings{li-liang-2021-prefix,
  title={Prefix-Tuning: Optimizing Continuous Prompts for Generation},
  author={Li, Xiang Lisa and Liang, Percy},
  booktitle={Proceedings of the 59th Annual Meeting of the Association for Computational Linguistics and the 11th International Joint Conference on Natural Language Processing (Volume 1: Long Papers)},
  pages={4582--4597},
  year={2021},
  doi={10.18653/v1/2021.acl-long.353},
  url={https://aclanthology.org/2021.acl-long.353/},
  publisher={Association for Computational Linguistics},
  address={Online}
}

@inproceedings{yang-klein-2021-fudge,
  title={{FUDGE}: Controlled Text Generation with Future Discriminators},
  author={Yang, Kevin and Klein, Dan},
  booktitle={Proceedings of the 2021 Conference of the North American Chapter of the Association for Computational Linguistics: Human Language Technologies},
  pages={3511--3535},
  year={2021}
}

@inproceedings{krause-etal-2021-gedi-generative,
  title={{GeDi}: Generative Discriminator Guided Sequence Generation},
  author={Krause, Ben and Gotmare, Akhilesh Deepak and McCann, Bryan and Keskar, Nitish Shirish and Joty, Shafiq and Socher, Richard and Rajani, Nazneen Fatema},
  booktitle={Findings of the Association for Computational Linguistics: EMNLP 2021},
  pages={4929--4952},
  year={2021}
}

@article{sotolar2024empo,
  title   = {{EmPO}: Emotion Grounding for Empathetic Response
             Generation through Preference Optimization},
  author  = {Sotolar, Ondrej and
             Formanek, Vojtech and
             Debnath, Alok and
             Lahnala, Allison and
             Welch, Charles and
             Flek, Lucie},
  journal = {arXiv preprint arXiv:2406.19071},
  year    = {2024}
}

@inproceedings{dathathri-etal-2020-plug-play,
  title={Plug and Play Language Models: A Simple Approach to Controlled Text Generation},
  author    = {Dathathri, Sumanth and
               Madotto, Andrea and
               Lan, Janice and
               Hung, Jane and
               Frank, Eric and
               Molino, Piero and
               Yosinski, Jason and
               Liu, Rosanne},
  booktitle={International Conference on Learning Representations},
  year      = {2020},
  url       = {https://openreview.net/forum?id=H1edEyBKDS}
}

@inproceedings{hu-etal-2024-aptness,
  author    = {Hu, Yuxuan and
               Tan, Minghuan and
               Zhang, Chenwei and
               Li, Zixuan and
               Liang, Xiaodan and
               Yang, Min and
               Li, Chengming and
               Hu, Xiping},
  title={{APTNESS}: Incorporating Appraisal Theory and Emotion Support Strategies for Empathetic Response Generation},
  booktitle={Proceedings of the 33rd ACM International Conference on Information and Knowledge Management},
  pages     = {900--909},
  year      = {2024},
  publisher={Association for Computing Machinery},
  doi={10.1145/3627673.3679687},
  url={https://doi.org/10.1145/3627673.3679687}
}

@inproceedings{pascual-etal-2021-plug-play,
  title={A Plug-and-Play Method for Controlled Text Generation},
  author    = {Pascual, Damian and
               Egressy, Beni and
               Meister, Clara and
               Cotterell, Ryan and
               Wattenhofer, Roger},
  editor    = {Moens, Marie-Francine and
               Huang, Xuanjing and
               Specia, Lucia and
               Yih, Scott Wen-tau},
  booktitle={Findings of the Association for Computational Linguistics: EMNLP 2021},
  month     = nov,
  year      = {2021},
  address={Punta Cana, Dominican Republic},
  publisher={Association for Computational Linguistics},
  pages     = {3973--3997},
  doi={10.18653/v1/2021.findings-emnlp.334},
  url={https://aclanthology.org/2021.findings-emnlp.334/}
}

@inproceedings{Zhang20:bertscore,
  title     = {{BERTScore}: Evaluating Text Generation with {BERT}},
  author    = {Tianyi Zhang and
               Varsha Kishore and
               Felix Wu and
               Kilian Q. Weinberger and
               Yoav Artzi},
  year      = {2020},
  booktitle = {Proceedings of the International Conference on Learning Representations}
}

@inproceedings{lin-2004-rouge,
  title = "{ROUGE}: A Package for Automatic Evaluation of Summaries",
  author = "Lin, Chin-Yew",
  booktitle = "Text Summarization Branches Out",
  month = jul,
  year = "2004",
  address = "Barcelona, Spain",
  publisher = "Association for Computational Linguistics",
  url = "https://aclanthology.org/W04-1013/",
  pages = "74--81"
}

@inproceedings{papineni-etal-2002-bleu,
  title = "{B}leu: a Method for Automatic Evaluation of Machine Translation",
  author = "Papineni, Kishore  and
    Roukos, Salim  and
    Ward, Todd  and
    Zhu, Wei-Jing",
  editor = "Isabelle, Pierre  and
    Charniak, Eugene  and
    Lin, Dekang",
  booktitle = "Proceedings of the 40th Annual Meeting of the Association for Computational Linguistics",
  month = jul,
  year = "2002",
  address = "Philadelphia, Pennsylvania, USA",
  publisher = "Association for Computational Linguistics",
  url = "https://aclanthology.org/P02-1040/",
  doi = "10.3115/1073083.1073135",
  pages = "311--318"
}

@inproceedings{banerjee-lavie-2005-meteor,
  title = "{METEOR}: An Automatic Metric for {MT} Evaluation with Improved Correlation with Human Judgments",
  author = "Banerjee, Satanjeev  and
    Lavie, Alon",
  editor = "Goldstein, Jade  and
    Lavie, Alon  and
    Lin, Chin-Yew  and
    Voss, Clare",
  booktitle = "Proceedings of the {ACL} Workshop on Intrinsic and Extrinsic Evaluation Measures for Machine Translation and/or Summarization",
  month = jun,
  year = "2005",
  address = "Ann Arbor, Michigan",
  publisher = "Association for Computational Linguistics",
  url = "https://aclanthology.org/W05-0909/",
  pages = "65--72"
}

@inproceedings{li-etal-2016-diversity,
  title = "A Diversity-Promoting Objective Function for Neural Conversation Models",
  author = "Li, Jiwei  and
    Galley, Michel  and
    Brockett, Chris  and
    Gao, Jianfeng  and
    Dolan, Bill",
  editor = "Knight, Kevin  and
    Nenkova, Ani  and
    Rambow, Owen",
  booktitle = "Proceedings of the 2016 Conference of the North {A}merican Chapter of the Association for Computational Linguistics: Human Language Technologies",
  month = jun,
  year = "2016",
  address = "San Diego, California",
  publisher = "Association for Computational Linguistics",
  url = "https://aclanthology.org/N16-1014/",
  doi = "10.18653/v1/N16-1014",
  pages = "110--119"
}

@article{jiang-etal-2023-mistral,
  author = {Jiang, Albert Q. and
            Sablayrolles, Alexandre and
            Mensch, Arthur and
            Bamford, Chris and
            Chaplot, Devendra Singh and
            de Las Casas, Diego and
            Bressand, Florian and
            Lengyel, Gianna and
            Lample, Guillaume and
            Saulnier, Lucile and
            Lavaud, L{\'e}lio Renard and
            Lachaux, Marie{-}Anne and
            Stock, Pierre and
            Le Scao, Teven and
            Lavril, Thibaut and
            Wang, Thomas and
            Lacroix, Timoth{\'e}e and
            El Sayed, William},
  title = {{Mistral 7B}},
  journal = {CoRR},
  volume = {abs/2310.06825},
  year = {2023},
  doi = {10.48550/arXiv.2310.06825},
  url = {https://arxiv.org/abs/2310.06825}
}

@inproceedings{he-etal-2023-debertav3,
  title = {{DeBERTaV3}: Improving {DeBERTa} using
           {ELECTRA}-Style Pre-Training with
           Gradient-Disentangled Embedding Sharing},
  author = {He, Pengcheng and
            Gao, Jianfeng and
            Chen, Weizhu},
  booktitle = {The Eleventh International Conference on
               Learning Representations},
  year = {2023},
  url = {https://openreview.net/forum?id=sE7-XhLxHA}
}

@article{traag-etal-2019-leiden,
  author = {Traag, V. A. and
            Waltman, L. and
            van Eck, N. J.},
  title = {From {Louvain} to {Leiden}: guaranteeing
           well-connected communities},
  journal = {Scientific Reports},
  volume = {9},
  pages = {5233},
  year = {2019},
  doi = {10.1038/s41598-019-41695-z},
  url = {https://doi.org/10.1038/s41598-019-41695-z}
}

@inproceedings{zhu-etal-2018-texygen,
  author = {Zhu, Yaoming and
            Lu, Sidi and
            Zheng, Lei and
            Guo, Jiaxian and
            Zhang, Weinan and
            Wang, Jun and
            Yu, Yong},
  editor = {Collins{-}Thompson, Kevyn and
            Mei, Qiaozhu and
            Davison, Brian D. and
            Liu, Yiqun and
            Yilmaz, Emine},
  title = {Texygen: {A} Benchmarking Platform for Text Generation Models},
  booktitle = {The 41st International {ACM} {SIGIR} Conference on
               Research {\&} Development in Information Retrieval,
               {SIGIR} 2018, Ann Arbor, MI, USA, July 08-12, 2018},
  pages = {1097--1100},
  publisher = {{ACM}},
  year = {2018},
  doi = {10.1145/3209978.3210080},
  url = {https://doi.org/10.1145/3209978.3210080}
}
\appendix
\section{Additional Related Work}
\label{app:additional_related}

\paragraph{Empathetic dialogue and emotional support.}
Empathetic dialogue generation has been widely studied as the problem of
recognizing an interlocutor's affective state and producing an appropriate
response. EmpatheticDialogues provides a representative benchmark for
emotionally grounded open-domain conversations
\citep{rashkin-etal-2019-towards}. Earlier and subsequent work has explored
explicit emotion conditioning, continuous affect representations, emotion
distributions, emotion mimicry, fine-grained emotional cues, user feedback, and
commonsense cognition for empathetic response generation
\citep{zhou-etal-2018-emotional,
colombo-etal-2019-affect-driven,
lin-etal-2019-moel,
majumder-etal-2020-mime,
li-etal-2020-empdg,
sabour-etal-2022-cem,
zhou-etal-2023-case}. Emotional support conversation further extends this line of work from one-shot empathy to multi-turn support, introducing support strategies, user states, strategy mixtures, turn-level transitions, initiative control, helpfulness optimization, and discourse dynamics \citep{liu-etal-2021-towards,tu-etal-2022-misc,zhao-etal-2023-transesc,deng-etal-2023-knowledge,li-etal-2024-helpful,wan-etal-2025-emodynamix,zhang-etal-2025-intentionesc}. Recent LLM-based systems also improve empathetic generation by injecting commonsense or future-aware inferences \citep{wang-etal-2025-sibyl}. These works motivate the importance of intermediate planning for affective dialogue, but they usually plan through predefined emotions, strategies, intentions, or external commonsense variables. Our work instead studies listener stance as the response-side orientation that mediates between dialogue context and surface realization.

\paragraph{Interactional stance and response-side uptake.}
Our notion of listener stance is related to work on interpersonal stancetaking, where conversational meaning is understood as positioning the speaker toward the interlocutor, the topic, and the ongoing interaction. Computational work has operationalized interactional stance through dimensions such as affect, investment, and alignment \citep{kiesling-etal-2018-interactional}. We focus on a response-side variant of this idea. Rather than modeling the previous speaker's private emotion, listener stance describes how the next speaker should take up the previous turn. This makes the variable role- and transition-sensitive: the same source utterance may invite different affective or interpersonal uptake depending on who is responding and how the local interaction has evolved. Prior emotional-support transition models capture related dynamics, but typically through discrete emotion, strategy, or state transitions \citep{zhao-etal-2023-transesc,wan-etal-2025-emodynamix}. EmoStance instead learns a latent stance space without requiring predefined stance names.

\paragraph{Ambiguity, disagreement, and soft supervision.}
Stance interpretation in dialogue is often subjective and underdetermined. The same utterance can reasonably invite multiple affective or interpersonal responses, and different annotators may emphasize different aspects of the local context. This connects our work to studies arguing that disagreement should be preserved rather than treated as annotation noise. Prior work has shown the limitations of collapsing subjective
judgments into a single hard label and has proposed methods for
learning from annotator distributions, disagreement patterns,
or individualized annotator behavior
\citep{fornaciari-etal-2021-beyond,
davani-etal-2022-dealing,
uma-etal-2021-learning,
wu-etal-2024-handling,
rodriguez-barroso-etal-2024-federated}.
Predictive uncertainty has also been used to measure label
ambiguity in subjective tasks
\citep{alies-etal-2025-measuring}.
EmojiDialogue follows this direction by preserving
multi-annotator emoji votes and confidence scores as soft
supervision rather than converting them into a single gold
stance label.

\paragraph{Emoji as affective and interpersonal signals.}
Emoji have been used as compact affective and semantic signals in representation learning and dialogue generation. Emoji2vec learns emoji embeddings from Unicode descriptions \citep{eisner-etal-2016-emoji2vec}, while DeepMoji uses large-scale emoji prediction to induce transferable representations for affective understanding \citep{felbo-etal-2017-using}. Recent work further shows that emoji sequences can express compositional meanings beyond one-to-one emotion labels \citep{yang-etal-2024-elco}. MojiTalk is especially relevant because it uses naturally occurring response emoji in Twitter conversations as emotional supervision and as control codes for response generation \citep{zhou-wang-2018-mojitalk}. Our use of emoji is different in three ways. First, emoji are used as weak supervision for inducing listener stance, not as the final prediction target. Second, we aggregate multiple emoji annotations into soft distributions instead of selecting a single discrete emoji label. Third, the learned stance representation controls a frozen instruction-tuned LLM through continuous prefix embeddings, rather than requiring the model to generate or condition on explicit emoji tokens at test time.

\paragraph{LLM-based annotation.}
Because EmojiDialogue uses multiple LLM annotators to obtain emoji-based weak supervision, it is also related to LLM-based data annotation. Recent studies show that LLMs can provide scalable annotations for subjective and social-science tasks, while also emphasizing the need to audit reliability, bias, and consistency \citep{gilardi2023chatgpt,tan-etal-2024-large}. Our setting uses LLM annotators not to create hard gold labels, but to obtain multiple weak signals whose disagreement and confidence structure are preserved. This design is intended to support scalable listener-stance supervision while avoiding the claim that any single emoji label is a human gold standard for emotion or mental state.

\paragraph{Controllable generation and continuous prompts.}
Controllable generation methods steer language models using discrete labels,
attribute classifiers, decoding-time discriminators, natural-language
instructions, or continuous prompts. PPLM steers generation by using gradients from attribute models to perturb a
pretrained language model's hidden activations
\citep{dathathri-etal-2020-plug-play}.
\citet{pascual-etal-2021-plug-play} propose a separate plug-and-play decoding
method that shifts the vocabulary distribution toward words semantically
related to a supplied topic or keyword. FUDGE and GeDi steer decoding with
future or generative discriminators
\citep{yang-klein-2021-fudge,krause-etal-2021-gedi-generative}.
Prefix-tuning keeps the base language model frozen and optimizes continuous
prefix vectors that act as virtual tokens \citep{li-liang-2021-prefix}.
Recent work further studies whether LLMs can smoothly control attribute
intensity through prompt-based interfaces \citep{zhou-etal-2024-evaluating}.
EmoStance follows the continuous-control direction, but its control signal is
not a manually specified attribute, a binary discriminator target, or a
natural-language instruction. Instead, the control vector is induced from emoji
weak supervision, predicted from the dialogue context and role transition, and
injected into a frozen generator through learned prefix embeddings.
\section{Dataset Annotation Details and Emoji Usage Statistics}
\label{app:dataset-details}

This appendix provides additional details for the construction and validation
of \textsc{EmojiDialogue}, including the source corpus processing,
adjacent-turn example construction, emoji inventory, human screening protocol,
LLM annotation prompt format, confidence statistics, emoji usage patterns, and
human plausibility audit. Artifact licenses, intended use, privacy checks, and
release conditions are discussed separately in
Appendix~\ref{app:reproducibility}.

\subsection{Source Corpus and Adjacent-Turn Example Construction}
\label{app:dataset-construction}

\textsc{EmojiDialogue} is built on top of EmpatheticDialogues
\citep{rashkin-etal-2019-towards}, an English dyadic dialogue corpus in which
each conversation is grounded in an emotional situation. EmpatheticDialogues
provides 32 situation-level emotion categories. We use these categories as part
of the original corpus context, but we do not treat them as utterance-level
emotion labels or as direct supervision targets for response generation.

The utterance-level annotation layer covers 99,556 utterances for each LLM
annotator. Since each utterance is annotated independently by four annotator
models, the full annotation layer contains four emoji judgments and four
confidence scores per utterance.

We construct source--response examples from adjacent dialogue turns. For a
turn pair \((u_t, u_{t+1})\), the source contains the situation description,
the dialogue history up to \(u_t\), and a marker indicating the next speaker.
The target response is \(u_{t+1}\). We split the corpus at the dialogue level
before constructing adjacent-turn examples, so that no dialogue contributes
turns to more than one partition. This prevents leakage of dialogue context
across the training, validation, and test splits.

\begin{table}[t]
\centering
\small
\begin{tabular}{lr}
\toprule
Split & Adjacent-turn examples \\
\midrule
Train & 58,829 \\
Validation & 9,263 \\
Test & 8,397 \\
\midrule
Total & 76,489 \\
\bottomrule
\end{tabular}
\caption{
Prepared adjacent-turn source--response examples after dialogue-level
splitting.
}
\label{tab:dialogue-split}
\end{table}

\subsection{Emoji Inventory and Human Screening}
\label{app:emoji-inventory}

We construct the initial emoji universe from the Python \texttt{emoji} package.
For reproducibility, we fix the Python \texttt{emoji} package version to
\texttt{0.1.0}. The extracted initial emoji universe contains 845 emoji
entries, denoted as \(\mathcal{E}_{\mathrm{pkg}}\).

Not all emoji in the original Python emoji universe are suitable as weak
affective, interpersonal, or conversational-attitude signals. We therefore
conduct a human screening step. Three volunteer screeners independently review
the initial emoji universe and vote on whether each emoji can plausibly express
an affective state, interpersonal stance, or conversational attitude.

The three screeners select 116, 122, and 107 emoji, respectively, from the
845-entry initial universe. Their unanimous intersection contains 96 emoji,
while their union contains 136 emoji. Since emoji-based affective--attitudinal
cues are inherently subjective, we use the union of the three screeners'
selections as the affective candidate pool. This choice allows the pool to
retain boundary cases and rare but potentially meaningful affective--attitudinal
signals that may be accepted by only one screener. Formally, the candidate pool
is defined as
\[
\begin{aligned}
\mathcal{E}_{\mathrm{cand}}
&= \mathcal{E}_{A} \cup \mathcal{E}_{B} \cup \mathcal{E}_{C}, \\
\mathcal{E}_{\mathrm{cand}}
&\subseteq \mathcal{E}_{\mathrm{pkg}}, \\
|\mathcal{E}_{\mathrm{cand}}|
&= 136.
\end{aligned}
\]

Rare but semantically meaningful affective emoji are retained at this stage so
that the candidate pool does not prematurely remove low-frequency but valid
affective--attitudinal cues. Sparsity is handled later by the downstream emoji
graph and affective-orientation projection components. In the final annotated
version of \textsc{EmojiDialogue}, 124 out of the 136 candidate emoji are
selected by at least one annotator model, indicating that most of the screened
candidate pool is used during annotation.

\begin{table}[t]
\centering
\small
\begin{tabular}{lr}
\toprule
Statistic & Value \\
\midrule
Initial emoji universe & 845 \\
Screener A selected & 116 \\
Screener B selected & 122 \\
Screener C selected & 107 \\
Selected by all three screeners & 96 \\
Selected by at least two screeners & 113 \\
Selected by at least one screener & 136 \\
Observed annotated emoji & 124 \\
\bottomrule
\end{tabular}
\caption{
Human screening statistics for affective emoji candidate selection.
Each screener independently judged whether an emoji could plausibly express an
affective state, interpersonal stance, or conversational attitude. The final
candidate pool is defined as the union of the three screeners' selections.
}
\label{tab:emoji-screening-agreement}
\end{table}

\subsection{LLM Annotation Protocol and Confidence-Weighted Soft Emoji Aggregation}
\label{app:llm-annotation-prompt}

Each utterance in \textsc{EmojiDialogue} is annotated independently by four LLM
annotators: DeepSeek-V3.2, Claude-Sonnet-4.6, Gemini-2.5-Pro, and GPT-5.4.
For each utterance, the annotator receives the situation description, the
dialogue context, the current speaker role, the current utterance, and the
screened 136-emoji candidate pool. The annotator is instructed to select
exactly one emoji from the candidate pool and to provide a confidence score on
a five-point scale. The original situation-level emotion category from
EmpatheticDialogues is not treated as an utterance-level supervision target.

The annotation prompt follows the format below.

\begin{quote}
\small
\textbf{Task.} Given the situation, dialogue context, current speaker role, and
current utterance, choose one emoji from the provided candidate emoji list. The
emoji should reflect the utterance's affective state, interpersonal stance, or
conversational attitude in context.

\textbf{Constraints.} Select exactly one emoji. Use only emoji from the provided
candidate list. Do not introduce new emoji outside the list.

\textbf{Confidence.} Provide a confidence score from 1 to 5, where higher
scores indicate higher confidence in the selected emoji.

\textbf{Output.} Return the selected emoji and the confidence score using the
specified structured output fields: \texttt{emoji} and \texttt{confidence}.
\end{quote}

The resulting annotations are used as weak affective--attitudinal observations,
not as gold emotion labels or gold listener-stance labels. For an utterance
\(u_i\), let \(a_{i,m} \in \mathcal{E}_{\mathrm{cand}}\) denote the emoji
selected by annotator \(m\), and let \(c_{i,m} \in \{1,\ldots,5\}\) denote the
corresponding confidence score, where \(m \in \{1,\ldots,4\}\). Rather than
collapsing the four annotations into a majority label, we aggregate them into
an utterance-level soft emoji distribution using confidence-normalized
annotator weights. Specifically, we define
\[
\alpha_{i,m}
=
\frac{c_{i,m}}
{\sum_{m'=1}^{4} c_{i,m'}},
\qquad
\sum_{m=1}^{4} \alpha_{i,m} = 1,
\]
and compute the soft emoji distribution as
\[
q_i^{E}(e)
=
\sum_{m=1}^{4}
\alpha_{i,m}
\mathbb{I}[a_{i,m}=e],
\qquad
e \in \mathcal{E}_{\mathrm{cand}}.
\]
Equivalently,
\[
q_i^{E}(e)
=
\frac{
\sum_{m=1}^{4}
c_{i,m}\mathbb{I}[a_{i,m}=e]
}{
\sum_{m=1}^{4} c_{i,m}
},
\qquad
e \in \mathcal{E}_{\mathrm{cand}}.
\]

This confidence-weighted representation preserves multi-annotator ambiguity
and disagreement while allowing higher-confidence annotations to contribute
more mass to the corresponding emoji. If all four annotators assign the same
confidence score, the formulation reduces to an unweighted vote distribution.
The raw confidence scores are also retained as annotation metadata and analyzed
descriptively in Appendix~\ref{app:confidence-statistics}.

\subsection{Dataset Format and Confidence Statistics}
\label{app:confidence-statistics}

Each dialogue-turn record in \textsc{EmojiDialogue} contains the situation
description, dialogue context, speaker role, current utterance, emoji
annotations selected by the four annotator models, and the corresponding
confidence scores. This structure allows the model to use both the textual
dialogue context and weak affective--attitudinal supervision signals from
multiple annotators.

Table~\ref{tab:emoji-confidence} reports the confidence-score distribution for
the four LLM annotators on the full utterance-level annotation set. Each
annotator contributes 99,556 single-utterance emoji annotations. Rows
corresponding to confidence levels with zero count are omitted.

\begin{table}[t]
\centering
\small
\begin{tabular}{lccc}
\toprule
\textbf{Annotator model} & \textbf{Confidence} & \textbf{Count} & \textbf{Proportion} \\
\midrule
DeepSeek-V3.2 & 2 & 22 & 0.0221\% \\
DeepSeek-V3.2 & 3 & 9,054 & 9.0944\% \\
DeepSeek-V3.2 & 4 & 67,542 & 67.8432\% \\
DeepSeek-V3.2 & 5 & 22,938 & 23.0403\% \\
Claude-Sonnet-4.6 & 1 & 2 & 0.0020\% \\
Claude-Sonnet-4.6 & 2 & 52 & 0.0522\% \\
Claude-Sonnet-4.6 & 3 & 1,652 & 1.6594\% \\
Claude-Sonnet-4.6 & 4 & 58,949 & 59.2119\% \\
Claude-Sonnet-4.6 & 5 & 38,901 & 39.0745\% \\
Gemini-2.5-Pro & 1 & 18 & 0.0181\% \\
Gemini-2.5-Pro & 2 & 45 & 0.0452\% \\
Gemini-2.5-Pro & 3 & 2,296 & 2.3062\% \\
Gemini-2.5-Pro & 4 & 41,529 & 41.7142\% \\
Gemini-2.5-Pro & 5 & 55,668 & 55.9163\% \\
GPT-5.4 & 1 & 44 & 0.0442\% \\
GPT-5.4 & 2 & 588 & 0.5906\% \\
GPT-5.4 & 3 & 12,057 & 12.1108\% \\
GPT-5.4 & 4 & 54,363 & 54.6054\% \\
GPT-5.4 & 5 & 32,504 & 32.6490\% \\
\bottomrule
\end{tabular}
\caption{
Confidence-score distribution of emoji annotations across the four LLM
annotators. Each annotator contributes 99,556 single-utterance annotations on
the full \textsc{EmojiDialogue} dataset.
}
\label{tab:emoji-confidence}
\end{table}

\subsection{Emoji Usage Statistics}
\label{app:emoji-usage}

This section reports emoji usage statistics for the full
\textsc{EmojiDialogue} dataset. All statistics are computed on the complete
dataset and do not distinguish between the training, validation, and test
splits. The affective emoji candidate pool contains 136 emoji types. Each
utterance is annotated once by each of the four annotator models:
DeepSeek-V3.2, Claude-Sonnet-4.6, Gemini-2.5-Pro, and GPT-5.4. Therefore,
each model contributes 99,556 single-utterance emoji annotations on the full
dataset.

\subsubsection{Frequency-Binned Emoji Usage by Annotator Model}
\label{app:emoji-usage-binned}

To summarize the overall shape of emoji usage, we group candidate emoji by
their usage frequency within each annotator model. Because the distribution is
highly skewed and long-tailed, we use the following frequency bins:
\(0\), \(1\text{--}10\), \(11\text{--}50\), \(51\text{--}100\),
\(101\text{--}500\), \(501\text{--}1{,}000\), \(1{,}001\text{--}5{,}000\),
and \(>5{,}000\). These bins provide a compact view of both the head and the
tail of the distribution while keeping the figure readable.

\begin{figure*}[t]
    \centering
    \includegraphics[width=\textwidth]{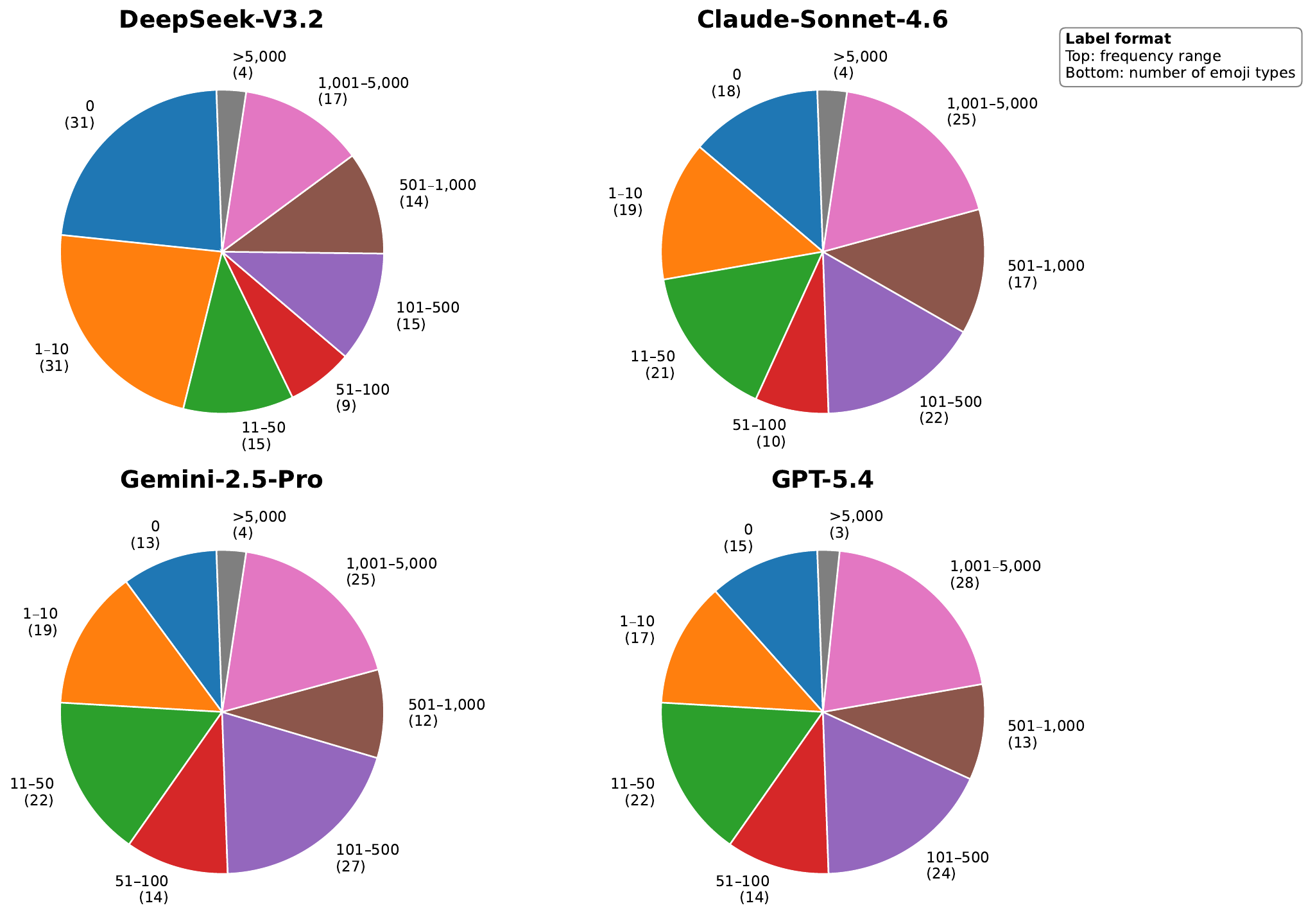}
    \caption{
    Frequency-binned emoji usage by annotator model on the full
    \textsc{EmojiDialogue} dataset. Each pie chart reports the number of emoji
    \emph{types} in each usage-frequency bin for one annotator model. Each
    model contributes \(N=99{,}556\) single-utterance annotations. The numbers
    of observed emoji types out of the 136-candidate pool are 105 for
    DeepSeek-V3.2, 118 for Claude-Sonnet-4.6, 123 for Gemini-2.5-Pro, and
    121 for GPT-5.4.
    }
    \label{fig:emoji-usage-pie}
\end{figure*}

Figure~\ref{fig:emoji-usage-pie} presents one pie chart for each annotator
model. Each slice denotes the number of emoji \emph{types} whose usage counts
fall into a given frequency bin. Thus, the pie charts summarize how the 136
candidate emoji are distributed across usage-frequency intervals, rather than
the raw number of utterance-level annotations themselves.

Overall, the four annotator models show a similar long-tail pattern. A
relatively small number of emoji occupy the high-frequency bins, while a large
portion of the candidate pool lies in low-frequency or zero-frequency bins.
DeepSeek-V3.2 leaves 31 emoji unused, Claude-Sonnet-4.6 leaves 18 unused,
Gemini-2.5-Pro leaves 13 unused, and GPT-5.4 leaves 15 unused. At the same
time, each model also uses a substantial number of medium- and high-frequency
emoji, indicating that the candidate pool is broad enough to support diverse
annotation behavior without forcing all candidate emoji to be used.

\subsubsection{Low-Frequency and Zero-Frequency Emoji}
\label{app:emoji-usage-lowfreq}

\begin{figure*}[t]
    \centering
    \includegraphics[width=\textwidth]{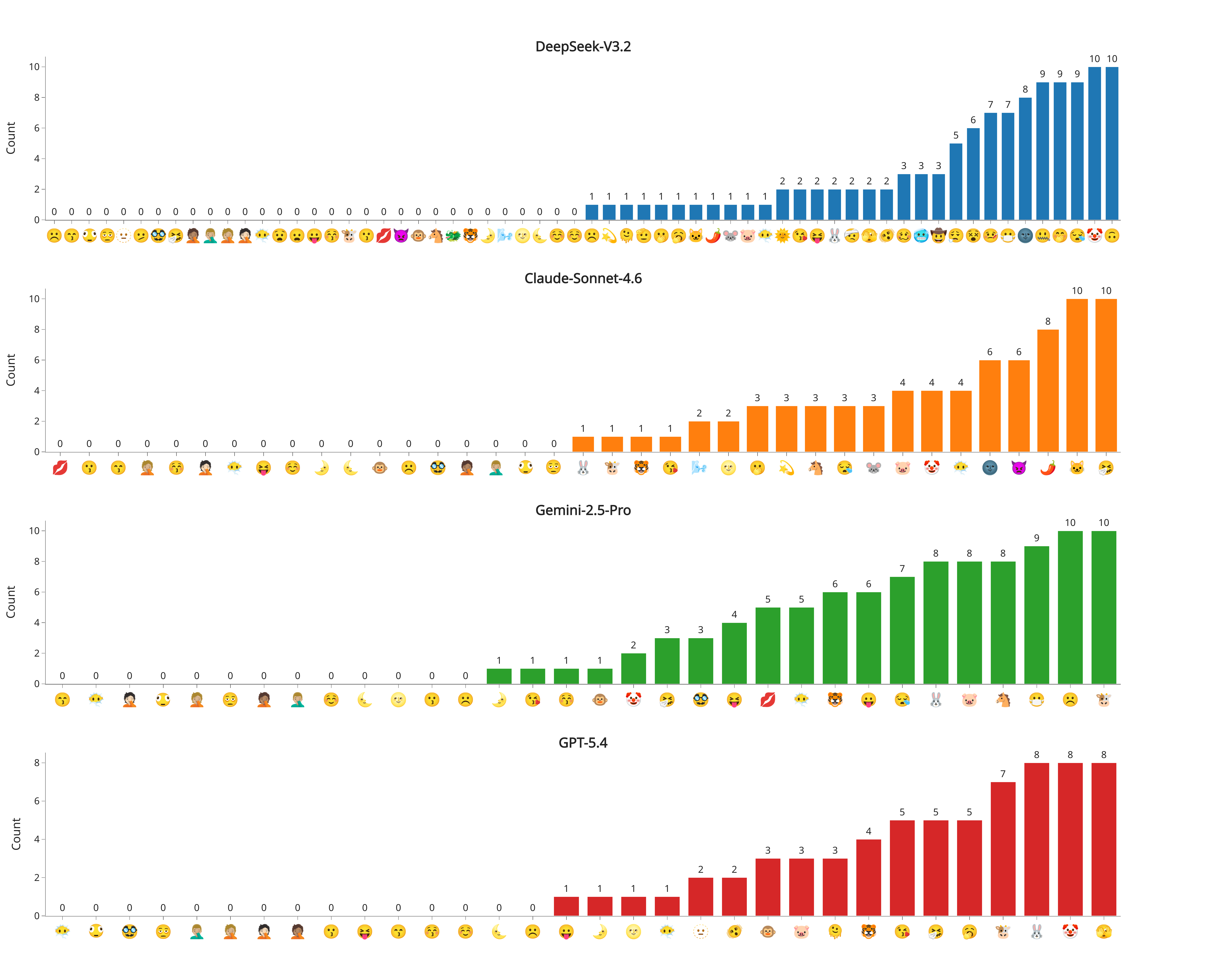}
    \caption{
    Tail usage of candidate emoji across annotator models. Each panel includes
    emoji with per-model counts no greater than 10, including zero-frequency
    candidates.
    }
    \label{fig:emoji-usage-lowfreq}
\end{figure*}

We next focus on the tail of the distribution by examining candidate emoji with
usage counts less than or equal to 10. This subset includes both low-frequency
emoji (\(1 \leq \text{count} \leq 10\)) and zero-frequency emoji
(\(\text{count}=0\)). Figure~\ref{fig:emoji-usage-lowfreq} visualizes these
emoji for each annotator model using vertical bar charts. The x-axis shows the
emoji symbols themselves, and the y-axis shows the corresponding usage counts.
Zero-frequency emoji are included explicitly and labeled with count 0.

This figure is useful for understanding which candidate emoji remain rarely
used or entirely unused by a given model. In particular, DeepSeek-V3.2 has 62
emoji with counts less than or equal to 10, including 31 zero-frequency emoji;
Claude-Sonnet-4.6 has 37 such emoji, including 18 zero-frequency emoji;
Gemini-2.5-Pro has 32, including 13 zero-frequency emoji; and GPT-5.4 also has
32, including 15 zero-frequency emoji. These results further support the view
that the candidate pool is not overly narrow: each annotator model uses only a
subset of the available inventory, while still leaving a visible long tail of
rare or unused emoji.

\subsection{Human Plausibility Audit of Emoji Weak Annotations}
\label{app:emoji-audit}

To evaluate the plausibility of the emoji weak annotations, we conduct a
representative dialogue-level human audit. The goal of this audit is to
determine whether the emoji assigned by an LLM annotator is plausible in the
full dialogue context, rather than to produce gold-standard emotion or
listener-stance labels.

We randomly sample 300 model-dialogue packages from the finalized
training/validation/test splits, proportional to their sizes: 231 training, 36
validation, and 33 test packages. Each package contains a complete dialogue
paired with one LLM annotator. The four LLM annotators are evenly represented,
with 75 packages per model.

Each dialogue is presented to human annotators one at a time. For each turn,
annotators see the original English text and the emoji assigned by the sampled
LLM. The model identity and confidence score are hidden. Annotators judge
whether the emoji plausibly expresses the utterance's affective state,
interpersonal stance, or conversational attitude in context. The available
labels are \textit{reasonable}, \textit{questionable but acceptable}, and
\textit{clearly unreasonable}.

Each turn is independently reviewed by three annotators. Judgments are
aggregated by majority vote: a turn is marked \textit{invalid} if at least two
annotators choose \textit{clearly unreasonable}, \textit{ambiguous} if at
least two choose \textit{questionable but acceptable} or if all three
annotators disagree, and \textit{valid} otherwise. Turn-level proportions are
reported with 95\% package-clustered bootstrap confidence intervals.

\begin{table*}[t]
\centering
\small
\begin{tabular}{lrrrrrr}
\toprule
Subset & Packages & Turns & Valid & Ambiguous & Invalid & Plausible \\
\midrule
Overall & 300 & 1297 & 99.69\% & 0.08\% & 0.23\% & 99.77\% \\
Claude-Sonnet-4.6 & 75 & 316 & 99.37\% & 0.00\% & 0.63\% & 99.37\% \\
DeepSeek-V3.2 & 75 & 334 & 100.00\% & 0.00\% & 0.00\% & 100.00\% \\
GPT-5.4 & 75 & 322 & 99.69\% & 0.31\% & 0.00\% & 100.00\% \\
Gemini-2.5-Pro & 75 & 325 & 99.69\% & 0.00\% & 0.31\% & 99.69\% \\
\bottomrule
\end{tabular}
\caption{
Human plausibility audit of emoji weak annotations. Each package contains a
complete dialogue paired with a hidden LLM annotator. Plausible is the sum of
Valid and Ambiguous. This audit evaluates weak-annotation plausibility in
context rather than gold-standard emotion or listener-stance correctness.
}
\label{tab:emoji-audit}
\end{table*}

The audit confirms that the weak annotations are highly plausible. Out of
1,297 turn-level items, 1,293 are judged valid, 1 ambiguous, and 3 invalid,
corresponding to a valid rate of 99.69\% with a 95\% confidence interval of
[99.38\%, 99.92\%]. The plausible rate, which counts both valid and ambiguous
annotations, is 99.77\% with a 95\% confidence interval of
[99.46\%, 100.00\%]. All four annotators achieve plausible rates above 99.3\%,
with DeepSeek-V3.2 showing no majority-invalid or majority-ambiguous turns.

Inter-annotator agreement is also high: exact three-annotator agreement is
95.53\%, and pairwise agreement is 97.02\%. Fleiss' \(\kappa\) is 0.068,
reflecting the highly skewed judgment distribution toward \textit{reasonable}
rather than poor agreement. Overall, the audit supports the use of
LLM-assigned emoji as weak affective--attitudinal supervision, while
maintaining the distinction between contextual plausibility validation and
gold-standard emotion or listener-stance annotation.
\subsection{Human--LLM Distributional Audit}
\label{app:human-llm-audit}

The preceding plausibility audit evaluates whether an individual
LLM-assigned emoji is acceptable in context, but it does not measure whether
aggregated LLM and human annotators produce similar distributions. We therefore
conduct a separate audit that directly estimates human--LLM divergence at both
the emoji-symbol level and the induced latent affective-orientation level. The
purpose is to quantify sample-specific divergence, not to establish that LLM
annotations are equivalent to human gold labels or culturally universal.

We begin with the 8,397 utterances in the \textsc{EmojiDialogue} test split.
Using a fixed seed, we stratify the candidates by the entropy of the existing
four-LLM confidence-weighted emoji distribution and sample 40 utterances from
each low-, medium-, and high-disagreement tertile. Each dialogue contributes at
most one utterance, yielding 120 unique utterances from 120 dialogues. Three
anonymous human annotators independently annotate every item using the same
136-emoji inventory and provide a 1--5 confidence score, producing 360 human
judgments. No new LLM annotations are requested. The emoji-to-region matrix
\(A\) is fixed before the audit, and no human annotation from this audit is used
to construct, select, or tune it.

\begin{table}[t]
\centering
\small
\begin{tabular}{lr}
\toprule
Audit item & Value \\
\midrule
Test candidate utterances & 8,397 \\
Sampled utterances & 120 \\
Low / medium / high disagreement & 40 / 40 / 40 \\
Unique dialogues & 120 \\
Emoji inventory & 136 \\
Observed emoji represented in \(A\) & 124 \\
Human annotators & 3 \\
Human judgments & 360 \\
\bottomrule
\end{tabular}
\caption{
Construction summary for the human--LLM emoji-distribution audit. The
remaining 12 inventory entries comprise 10 canonicalized variants and two
unselected zero-mass emoji.
}
\label{tab:human-llm-summary}
\end{table}

For each item, we compare the pre-existing confidence-weighted four-LLM emoji
distribution with the aggregated human distribution. The primary human
distribution uses unweighted votes, while confidence-weighted aggregation is
included as a robustness analysis. We evaluate both the original emoji-symbol
distributions and their projections through the fixed matrix \(A\) into the
nine-region affective-orientation space. Jensen--Shannon divergence uses
base-2 logarithms. Distributional overlap is defined as
\[
\operatorname{Overlap}(p,q)=\sum_i \min\{p_i,q_i\}.
\]
All confidence intervals are 95\% paired sample-level bootstrap intervals with
10,000 resamples.

\begin{table}[t]
\centering
\small
\setlength{\tabcolsep}{3pt}
\begin{tabular}{lcc}
\toprule
\shortstack[l]{Representation}
& \shortstack{JSD, mean\\{[95\% CI]}}
& \shortstack{Overlap, mean\\{[95\% CI]}} \\
\midrule
\shortstack[l]{Emoji-symbol\\level}
& \shortstack{0.442\\{[0.403, 0.482]}}
& \shortstack{0.449\\{[0.415, 0.483]}} \\
\shortstack[l]{Latent-region\\level}
& \shortstack{0.206\\{[0.173, 0.239]}}
& \shortstack{0.670\\{[0.630, 0.710]}} \\
\bottomrule
\end{tabular}
\caption{
Distributional comparison between aggregated LLM and human annotations before
and after projection through the fixed learned membership matrix \(A\).
}
\label{tab:human-llm-main}
\end{table}

The audit reveals non-trivial disagreement at the exact emoji-symbol level.
After projection, the paired change defined as region-level JSD minus
emoji-level JSD is \(-0.237\), with a 95\% interval of
\([-0.268,-0.206]\), while mean overlap increases from 0.449 to 0.670.
This pattern is consistent with symbol-specific variation: humans and LLMs may
select different emoji that the learned mapping associates with similar
affective or interpersonal orientations.

\begin{table}[t]
\centering
\small
\setlength{\tabcolsep}{3pt}
\begin{tabular}{lrrrr}
\toprule
\shortstack[l]{LLM-disagreement\\stratum}
& \(N\)
& \shortstack{Emoji\\JSD}
& \shortstack{Region\\JSD}
& \shortstack{Region\\overlap} \\
\midrule
Low    & 40 & 0.364 & 0.174 & 0.709 \\
Medium & 40 & 0.470 & 0.236 & 0.634 \\
High   & 40 & 0.493 & 0.207 & 0.667 \\
\bottomrule
\end{tabular}
\caption{
Human--LLM distributional divergence across strata defined by the entropy of
the original four-LLM distribution.
}
\label{tab:human-llm-strata}
\end{table}

The reduction occurs in all three disagreement strata. However, any shared
coarse projection can mechanically contract divergence. To test whether the
learned correspondence between emoji and regions is more meaningful than
arbitrary nine-region coarsening, we perform 1,000 row permutations of \(A\).
Each permutation preserves the set and shape of the soft membership rows while
randomly breaking their correspondence with emoji symbols.

\begin{table}[t]
\centering
\small
\setlength{\tabcolsep}{3pt}
\begin{tabular}{lr}
\toprule
Mapping
& \shortstack{Mean region-level\\JSD} \\
\midrule
Learned \(A\)
& 0.205526 \\
\shortstack[l]{Randomly permuted\\mappings, mean}
& 0.315756 \\
\shortstack[l]{Random-permutation\\2.5th percentile}
& 0.288704 \\
\bottomrule
\end{tabular}
\caption{
Comparison of the learned emoji-to-region mapping with 1,000 row-permuted
mappings. None of the permutations attains a JSD as low as the learned
mapping, giving a smoothed one-sided empirical value of \(p=.001\).
}
\label{tab:human-llm-permutation}
\end{table}

The learned mapping produces a lower region-level JSD than every permuted
mapping. Thus, the observed reduction is not explained only by projecting the
emoji into nine regions; the learned matrix groups some human--LLM
symbol-level disagreements more meaningfully than arbitrary assignments.

\begin{table*}[t]
\centering
\small
\begin{tabular}{lrrrr}
\toprule
Human aggregation
& Emoji JSD & Region JSD & Emoji overlap & Region overlap \\
\midrule
Unweighted votes
& 0.442 & 0.206 & 0.449 & 0.670 \\
Confidence-weighted votes
& 0.432 & 0.200 & 0.458 & 0.676 \\
\bottomrule
\end{tabular}
\caption{
Robustness of the human--LLM comparison to confidence-weighted aggregation of
the three human annotations.
}
\label{tab:human-llm-robustness}
\end{table*}

The result is robust to confidence-weighted aggregation of the human
annotations. Exact emoji selection is also highly variable among the human
annotators themselves.

\begin{table}[t]
\centering
\small
\begin{tabular}{lrr}
\toprule
Human exact-emoji pattern & Count & Percentage \\
\midrule
All three selected the same emoji & 0 & 0.0\% \\
Two selected the same emoji & 39 & 32.5\% \\
All three selected different emoji & 81 & 67.5\% \\
\bottomrule
\end{tabular}
\caption{
Exact-emoji agreement patterns among the three human annotators.
}
\label{tab:human-exact-emoji}
\end{table}

As contextual evidence for task subjectivity, a leave-one-human-out comparison
gives an emoji-level JSD of 0.851 and a region-level JSD of 0.465. This is not
a strictly matched baseline because it compares one human with the other two,
whereas the primary audit compares aggregated four-LLM and three-human
distributions.

We therefore do not treat the human annotations as a unique hard ground truth.
The audit provides a direct estimate of LLM--human divergence on a stratified
English test subset and shows that the learned region mapping is less sensitive
to exact-symbol disagreement than the original emoji space. It does not
establish equivalence to human annotation, cultural universality, or a complete
taxonomy of response-side affective orientation.
\section{Additional Method Details}
\label{app:method-details}

This appendix provides implementation details omitted from the main method
section, including role-marked serialization, weak emoji distribution
construction, name-free emoji membership construction, continuous
emoji-derived affective vectors, role-aware orientation transition priors, loss
computation, prefix projection, and candidate reranking. All emoji-derived
quantities used by the model are constructed from the training data. At
inference time, the model uses only the situation description, dialogue
history, and speaker-role markers.

Throughout this appendix, we write distributions as row vectors when
multiplying by the emoji-to-region membership matrix \(A\). The superscript
\(Z\) denotes latent emoji-induced affective-orientation regions, not a gold
listener-stance label space.

\subsection{Role-Marked Serialization}
\label{app:serialization}

For each adjacent-turn example \((u_t,u_{t+1})\), the serialized input contains
the optional situation description, the dialogue history up to turn \(t\), and
the role marker of the next speaker:
\[
\begin{split}
x_t =
[\mathrm{SIT}]~s~
[\mathrm{CTX}]~
\langle r_1\rangle u_1~
\cdots~
\\
\langle r_t\rangle u_t~
[\mathrm{NEXT}]~
\langle r_{t+1}\rangle .
\end{split}
\]
If no situation description is available, the situation segment is omitted.
The target response \(u_{t+1}\), emoji annotations, emotion labels, and all
weak affective-orientation targets are excluded from the input. The ordered
role transition is
\[
\rho_t = r_t \rightarrow r_{t+1}.
\]

\subsection{Weak Emoji Distributions}
\label{app:emoji-weak-distribution}

Let \(\mathcal{E}\) denote the screened emoji inventory. Suppose utterance
\(u_t\) receives \(n_t\) emoji judgments. Each judgment provides an emoji
\(e_{t,m}\in\mathcal{E}\) and, when available, a confidence weight
\(\alpha_{t,m}\). We aggregate the judgments into a soft emoji distribution:
\[
q_t^E(e)
=
\frac{
\sum_{m=1}^{n_t}
\alpha_{t,m}
\mathbb{I}[e_{t,m}=e]
}{
\sum_{m=1}^{n_t}
\alpha_{t,m}
+\epsilon
}.
\]
If confidence scores are not provided, we set \(\alpha_{t,m}=1\). Emoji outside
the screened inventory are discarded, and the remaining distribution is
renormalized. This soft distribution is used as the starting point for all
emoji-derived weak affective-orientation targets.

The soft distribution is important because the target object is not a
single-objective emotion category. For affective and interpersonal meanings,
annotator disagreement may indicate ambiguity or multiple plausible readings
rather than noise alone.

\subsection{Name-Free Emoji Membership Construction}
\label{app:emoji-graph}

We construct a name-free emoji membership matrix using only data-internal
relations among emoji. The construction combines contextual usage similarity
and annotator co-selection similarity. Emoji names are not used.

Let \(\psi(u_t)\) be a frozen utterance representation. The contextual centroid
of emoji \(e\) is
\[
\mathbf{c}_e
=
\frac{
\sum_t q_t^E(e)\psi(u_t)
}{
\sum_t q_t^E(e)+\epsilon
}.
\]
The contextual similarity between emoji \(e\) and \(e'\) is computed as
\[
S_{\mathrm{ctx}}(e,e')
=
\frac{
1+
\operatorname{cos}(\mathbf{c}_e,\mathbf{c}_{e'})
}{2}.
\]
The affine transformation maps cosine similarity into \([0,1]\).

For annotator co-selection similarity, we compare how often two emoji receive
mass on the same utterances:
\[
S_{\mathrm{conf}}(e,e')
=
\frac{
\sum_t q_t^E(e)q_t^E(e')
}{
\sqrt{\sum_t q_t^E(e)^2}
\sqrt{\sum_t q_t^E(e')^2}
+\epsilon
}.
\]
The final emoji affinity matrix is
\[
W
=
\frac{\lambda_{\mathrm{ctx}}}
{\lambda_{\mathrm{ctx}}+\lambda_{\mathrm{conf}}}
S_{\mathrm{ctx}}
+
\frac{\lambda_{\mathrm{conf}}}
{\lambda_{\mathrm{ctx}}+\lambda_{\mathrm{conf}}}
S_{\mathrm{conf}} .
\]
We sparsify $W$ by retaining the top-$k$ neighbors of
each emoji and then symmetrize the graph. We then apply
the Leiden community-detection algorithm
\citep{traag-etal-2019-leiden} to the sparse graph,
producing $K$ latent affective-orientation regions
$\mathcal{C} = \{C_1, \ldots, C_K\}$.

The emoji-to-region membership matrix
\[
A\in[0,1]^{|\mathcal{E}|\times K}
\]
maps emoji distributions into the latent affective-orientation space. Each row
of \(A\) sums to one:
\[
A_{e,k}\geq 0,
\qquad
\sum_{k=1}^{K}A_{e,k}=1 .
\]
For emoji that clearly belong to a single community, the membership is nearly
one-hot. For boundary emoji, we allow soft multi-region membership based on
their affinity to neighboring regions.

One implementation is to compute the affinity of emoji \(e\) to region \(k\)
as
\[
B_{e,k}
=
\frac{
\sum_{e'\in\mathcal{C}_k} W_{e,e'}
}{
\sum_{m=1}^{K}
\sum_{e'\in\mathcal{C}_m} W_{e,e'}
+\epsilon
},
\]
and then combine this soft affinity with the hard community assignment:
\[
A_{e,k}
=
(1-\delta)\mathbb{I}[e\in\mathcal{C}_k]
+
\delta B_{e,k},
\]
where \(\delta\in[0,1]\) controls the amount of boundary smoothing. The
utterance-level latent-region distribution is
\[
q_t^Z = q_t^E A .
\]
Since \(q_t^E\) is a probability distribution and each row of \(A\) is
normalized, \(q_t^Z\) is also a probability distribution over latent
affective-orientation regions.

\subsection{Continuous Emoji-Derived Vectors and Orientation Prototypes}
\label{app:emoji-centroids}

Each emoji \(e\) is assigned a continuous vector \(\mathbf{h}_e\) based on its
contextual usage in the training corpus:
\[
\mathbf{h}_e
=
\frac{
\sum_t q_t^E(e)\psi(u_t)
}{
\sum_t q_t^E(e)+\epsilon
}.
\]
The utterance-level continuous emoji-derived affective vector is the
emoji-weighted average
\[
v_t
=
\sum_{e\in\mathcal{E}}
q_t^E(e)\mathbf{h}_e .
\]
This vector preserves fine-grained information from the original emoji
distribution, including within-region variation.

For each latent affective-orientation region \(k\), we compute an orientation
prototype vector from the training set:
\[
\mu_k
=
\frac{
\sum_{t\in\mathcal{T}_{\mathrm{train}}}
q_t^Z(k)v_t
}{
\sum_{t\in\mathcal{T}_{\mathrm{train}}}
q_t^Z(k)
+\epsilon
},
\]
where \(\mathcal{T}_{\mathrm{train}}\) indexes training utterances. These
prototypes define the reconstruction map from a predicted orientation
distribution to a continuous control vector:
\[
\hat{v}
=
\sum_{k=1}^{K}
\hat{q}^Z(k)\mu_k .
\]
For next-response generation, this gives
\[
\hat{v}_{t+1}
=
\sum_{k=1}^{K}
\hat{q}_{t+1}^Z(k)\mu_k .
\]

\subsection{Role-Aware Response-Orientation Predictor}
\label{app:role-aware-details}

The orientation predictor uses an encoder to obtain a contextual
representation:
\[
\mathbf{h}_t=\operatorname{Enc}_{\theta}(x_t).
\]
A source-expression head predicts the source-side affective-expression
distribution of the latest observed utterance:
\[
a_t^{\mathrm{cur}}
=
f_{\mathrm{cur}}(\mathbf{h}_t),
\qquad
\hat{q}_t^Z
=
\operatorname{softmax}(a_t^{\mathrm{cur}}).
\]
We reconstruct an auxiliary source-side expression vector as
\[
\hat{v}_t
=
\sum_{k=1}^{K}
\hat{q}_t^Z(k)\mu_k .
\]
The compact source-side expression summary used by the response-orientation
head is
\[
d_t
=
\operatorname{MLP}_{d}
\left(
[
\hat{q}_t^Z;
\hat{v}_t
]
\right).
\]

Let \(\mathbf{e}_{\rho_t}\) be a learned embedding of the ordered role
transition \(\rho_t=r_t\rightarrow r_{t+1}\). The prior-free next-response
logits are
\[
\ell_{t+1}^{0}
=
f_{\mathrm{next}}
\left(
[
\mathbf{h}_t;
\mathbf{e}_{\rho_t};
d_t
]
\right).
\]
This head provides the neural estimate of the response-side affective
orientation before the role-aware transition prior is added.

To make the transition prior sensitive to source-expression uncertainty, we
compute the normalized entropy of the predicted source-side distribution:
\[
\bar{H}_t
=
-
\frac{1}{\log K}
\sum_{k=1}^{K}
\hat{q}_t^Z(k)
\log
\left(
\hat{q}_t^Z(k)+\epsilon
\right).
\]
The uncertainty-aware gate is
\[
\gamma_t
=
(1-\bar{H}_t)
\cdot
\sigma
\left(
\operatorname{MLP}_{g}
\left(
[
\mathbf{h}_t;
\mathbf{e}_{\rho_t};
d_t
]
\right)
\right),
\]
where \(\sigma(\cdot)\) is the sigmoid function. The gate reduces the influence
of the transition prior when the source-expression prediction is highly
uncertain.

\subsection{Role-Aware Transition Prior}
\label{app:transition-prior}

For each ordered role transition \(\rho\), we estimate a smoothed transition
matrix
\[
T^{\rho}\in\mathbb{R}^{K\times K}
\]
from the training set. The soft count from source-side affective-expression
region \(k\) to response-side affective-orientation region \(k'\) is
\[
N_{k,k'}^{\rho}
=
\sum_{t:\rho_t=\rho}
q_t^Z(k)
q_{t+1}^Z(k').
\]
With additive smoothing coefficient \(\alpha\), the transition probability is
\[
T_{k,k'}^{\rho}
=
\frac{
N_{k,k'}^{\rho}+\alpha
}{
\sum_{m=1}^{K}
\left(
N_{k,m}^{\rho}+\alpha
\right)
}.
\]

Given the predicted source-side affective-expression distribution
\(\hat{q}_t^Z\), the role-conditioned response-orientation prior is
\[
\pi_{t+1}(k')
=
\sum_{k=1}^{K}
\hat{q}_t^Z(k)
T_{k,k'}^{\rho_t}.
\]
Equivalently, writing distributions as row vectors,
\[
\pi_{t+1}
=
\hat{q}_t^Z T^{\rho_t}.
\]
The final next-response logits are
\[
\ell_{t+1}
=
\ell_{t+1}^{0}
+
\lambda_{\mathrm{tr}}
\gamma_t
\log(\pi_{t+1}+\epsilon),
\]
and the predicted response-orientation distribution is
\[
\hat{q}_{t+1}^Z
=
\operatorname{softmax}
(
\ell_{t+1}
).
\]

\subsection{Loss Details}
\label{app:loss-details}

The orientation predictor is trained with soft weak targets. For a weak target
distribution \(q\) and predicted distribution \(\hat{q}\), the soft
cross-entropy is
\[
\operatorname{CE}(q,\hat{q})
=
-
\sum_{k=1}^{K}
q(k)
\log
\left(
\hat{q}(k)+\epsilon
\right).
\]

To handle latent-region imbalance on the response-orientation side, we
optionally use weighted soft cross-entropy:
\[
\operatorname{CE}_{w}(q,\hat{q})
=
-
\sum_{k=1}^{K}
w_k q(k)
\log
\left(
\hat{q}(k)+\epsilon
\right).
\]
The class weight \(w_k\) is computed from the training frequency \(\varphi_k\)
of latent region \(k\):
\[
\varphi_k
=
\frac{
\sum_{t\in\mathcal{T}_{\mathrm{train}}}
q_{t+1}^Z(k)
}{
\sum_{m=1}^{K}
\sum_{t\in\mathcal{T}_{\mathrm{train}}}
q_{t+1}^Z(m)
+\epsilon
},
\]
\[
w_k
=
\left(
\frac{1}{\varphi_k+\epsilon}
\right)^{\beta},
\]
and the weights are normalized so that their mean is one:
\[
w_k
\leftarrow
\frac{
K w_k
}{
\sum_{m=1}^{K}w_m+\epsilon
}.
\]
The exponent \(\beta\) controls the strength of imbalance correction.

For example \((u_t,u_{t+1})\), the source-expression loss is
\[
L_t^{\mathrm{cur}}
=
\operatorname{CE}
\left(
q_t^Z,
\hat{q}_t^Z
\right).
\]
The response-orientation loss is
\[
L_t^{\mathrm{next}}
=
\operatorname{CE}_{w}
\left(
q_{t+1}^Z,
\hat{q}_{t+1}^Z
\right).
\]
We also use an auxiliary prior-free response-orientation loss:
\[
\hat{q}_{t+1}^{Z,0}
=
\operatorname{softmax}(\ell_{t+1}^{0}),
\]
\[
L_t^{0}
=
\operatorname{CE}_{w}
\left(
q_{t+1}^Z,
\hat{q}_{t+1}^{Z,0}
\right).
\]
This term applies the same weak supervision before the transition prior is
added, which stabilizes response-orientation prediction. If this auxiliary
term is not used, its coefficient can be set to zero.

The vector reconstruction loss is
\[
L_t^{\mathrm{vec}}
=
\left\|
\hat{v}_t-v_t
\right\|_2^2
+
\left\|
\hat{v}_{t+1}-v_{t+1}
\right\|_2^2,
\]
where
\[
\hat{v}_{t+1}
=
\sum_{k=1}^{K}
\hat{q}_{t+1}^Z(k)\mu_k .
\]
The full orientation-prediction objective is
\[
\begin{split}
&\mathcal{L}_{\mathrm{orient}}
=
\frac{1}{N}
\sum_{t=1}^{N}
\\
&\left(
\lambda_{\mathrm{next}}L_t^{\mathrm{next}}
+
\lambda_{0}L_t^{0}
+
\lambda_{\mathrm{cur}}L_t^{\mathrm{cur}}
+
\lambda_{\mathrm{vec}}L_t^{\mathrm{vec}}
\right).
\end{split}
\]
In the main text, this objective is written in simplified form to emphasize
the response-orientation prediction term, the vector reconstruction term, and
the auxiliary source-expression term.

\subsection{Prefix Projector}
\label{app:prefix-projector}

The frozen generator has embedding dimension \(d_{\Omega}\). The prefix
projector maps an orientation vector \(v\in\mathbb{R}^{d}\) into \(m\)
continuous prefix embeddings:
\[
R_{\omega}(v)
\in
\mathbb{R}^{m\times d_{\Omega}}.
\]
In our implementation, \(R_{\omega}\) is a lightweight MLP:
\[
R_{\omega}(v)
=
\operatorname{reshape}
\left(
W_2
\sigma(W_1v+b_1)
+
b_2
\right),
\]
where the output is reshaped into \(m\) prefix tokens. These prefix embeddings
are prepended to the token embeddings of the serialized dialogue context.

The generator parameters \(\Omega\) remain frozen, and only the projector
parameters \(\omega\) are updated with
\[
\begin{split}
&\mathcal{L}_{\mathrm{gen}}
=
\\
&-
\sum_{t}
\sum_{j=1}^{|u_{t+1}|}
\log
p_{\Omega}
\left(
u_{t+1,j}
\mid
P_{t+1},
x_t,
u_{t+1,<j}
\right).
\end{split}
\]
The loss is computed only over the response tokens. During projector training,
the weak response-side orientation vector \(v_{t+1}\) derived from the
observed response is used, with
\[
P_{t+1}=R_{\omega}(v_{t+1}).
\]
During inference, \(v_{t+1}\) is replaced by the reconstructed predicted vector
\(\hat{v}_{t+1}\).

\subsection{Candidate Reranking}
\label{app:reranking}

At decoding time, we optionally sample \(B\) candidate responses
\[
\left\{
\widetilde{u}_{t+1}^{(1)},
\ldots,
\widetilde{u}_{t+1}^{(B)}
\right\}
\]
from the same predicted response-side orientation vector \(\hat{v}_{t+1}\). To
score a candidate, we append it to the dialogue history:
\[
\widetilde{D}_{t}^{(b)}
=
D_{\leq t}
\cup
\left\{
(r_{t+1},\widetilde{u}_{t+1}^{(b)})
\right\}.
\]
The appended context is serialized and passed through the orientation scorer:
\[
\widetilde{x}_{t}^{(b)}
=
\operatorname{ser}
\left(
s,
\widetilde{D}_{t}^{(b)},
r_t
\right).
\]
The source-expression head is then used to estimate the orientation realized by
the appended candidate, which is now the latest observed turn:
\[
\widetilde{q}_{t+1}^{Z,(b)}
=
\operatorname{softmax}
\left(
f_{\mathrm{cur}}
\left(
\operatorname{Enc}_{\theta}
\left(
\widetilde{x}_{t}^{(b)}
\right)
\right)
\right).
\]

Each candidate is scored by its consistency with the intended response-side
orientation:
\[
J^{(b)}
=
D
\left(
\hat{q}_{t+1}^Z,
\widetilde{q}_{t+1}^{Z,(b)}
\right)
+
\eta
\mathcal{R}
\left(
\widetilde{u}_{t+1}^{(b)}
\right).
\]
By default, we use cross-entropy as the distributional divergence:
\[
\begin{split}
&D
\left(
\hat{q}_{t+1}^Z,
\widetilde{q}_{t+1}^{Z,(b)}
\right)
=
\\
&-
\sum_{k=1}^{K}
\hat{q}_{t+1}^Z(k)
\log
\left(
\widetilde{q}_{t+1}^{Z,(b)}(k)+\epsilon
\right).
\end{split}
\]

The length regularizer is optional. When used, it penalizes candidates whose
length deviates substantially from the expected response length:
\[
\mathcal{R}_{\mathrm{len}}(\widetilde{u})
=
\left(
\frac{
|\widetilde{u}|-\mu_{\ell}
}{
\sigma_{\ell}+\epsilon
}
\right)^2,
\]
where \(\mu_{\ell}\) and \(\sigma_{\ell}\) are estimated from training
responses.

The final response is
\[
b^*
=
\arg\min_{1\leq b\leq B}
J^{(b)},
\qquad
\hat{u}_{t+1}
=
\widetilde{u}_{t+1}^{(b^*)}.
\]
Reranking is applied only at decoding time and does not update any model
parameters.

\subsection{Training and Inference Protocol}
\label{app:training-inference-protocol}

Training consists of three preparation and optimization steps. First, weak
emoji distributions \(q_t^E\), the name-free membership matrix \(A\), latent
affective-orientation distributions \(q_t^Z\), continuous emoji-derived
affective vectors \(v_t\), orientation prototypes \(\mu_k\), and role-aware
transition matrices \(T^\rho\) are constructed from the training data. Second,
the role-aware orientation predictor is trained with
\(\mathcal{L}_{\mathrm{orient}}\). Third, the generator is kept frozen and the
prefix projector is trained with \(\mathcal{L}_{\mathrm{gen}}\) using weak
response-side orientation vectors derived from observed responses.

At inference time, emoji annotations are unavailable and are not required. For
each input \(x_t\), \textsc{EmoStance} predicts the source-side affective
expression \(\hat{q}_t^Z\), constructs the role-conditioned transition prior
\(\pi_{t+1}\), predicts the response-side affective orientation
\(\hat{q}_{t+1}^Z\), reconstructs the continuous control vector
\(\hat{v}_{t+1}\), maps it into prefix embeddings, and generates the response
with the frozen language model. Optional reranking can then be applied to
improve orientation consistency.

This protocol ensures that emoji annotations are used only as weak supervision
during training. They are not appended to the test-time input, are not treated
as gold emotion labels or gold listener-stance labels, and are not the desired
output of the system.
\section{Experimental Details and Supplementary Results}
\label{app:experiments}

\subsection{Data and Inference Setting}
\label{app:data-inference}

Table~\ref{tab:data-statistics} summarizes the prepared
\textsc{EmojiDialogue} split used for response-orientation prediction and
generation-control ablations.

\begin{table}[t]
\centering
\small
\begin{tabular}{lr}
\toprule
Statistic & Value \\
\midrule
Adjacent-turn examples & 76,489 \\
Training examples & 58,829 \\
Validation examples & 9,263 \\
Test examples & 8,397 \\
Observed emoji & 124 \\
Latent affective-orientation regions & 9 \\
Continuous orientation dimension & 256 \\
\bottomrule
\end{tabular}
\caption{
Statistics of the prepared \textsc{EmojiDialogue} split used for
response-orientation prediction and ablation experiments.
}
\label{tab:data-statistics}
\end{table}

Across all deployable settings, the model input is restricted to
inference-time information: the situation description, dialogue history up to
the current turn, and speaker-role markers. Emoji annotations, latent-region
targets, response-derived orientation vectors, and reference responses are
never provided at inference time. System-level automatic comparison with prior
empathetic-response systems is conducted on the full ED test set, where the
aligned evaluation set contains 5,255 examples.

\subsection{Baseline Details}
\label{app:baseline-details}

Table~\ref{tab:baseline-groups} summarizes the baseline groups used in the
main comparison. All baseline outputs are produced by our own reproduction
under the same aligned EmpatheticDialogues evaluation setting. All systems use
\texttt{mistralai/Mistral-7B-Instruct-v0.3} as the base generator and are
evaluated on the same aligned ED test contexts. The input contains only the
situation description, dialogue history, and speaker-role markers available at
inference time. No system receives reference responses, emoji annotations,
latent-region targets, or response-derived orientation vectors at test time.

\begin{table*}[t]
\centering
\small
\begin{tabular}{p{0.16\textwidth}p{0.21\textwidth}p{0.56\textwidth}}
\toprule
Baseline & Category & Purpose \\
\midrule
LLM-only
& Backbone generator
& Base instruction-tuned generator without affective control. \\

LLM-prompt
& Prompt-level control
& Uses verbal affective instructions only. \\

LLM-SFT
& Supervised tuning
& Supervised response learning without the latent affective-orientation
module. \\

EmPO-DPO
& Preference optimization
& Tests whether preference optimization alone explains the improvements. \\

CASE
& Prior ED system
& ED-compatible same-backbone adaptation of CASE
\citep{zhou-etal-2023-case}. \\

APTNESS
& Prior ED system
& ED-compatible same-backbone adaptation of APTNESS
\citep{hu-etal-2024-aptness}. \\

Sibyl
& Future-aware commonsense
& ED-compatible commonsense-augmented adaptation of Sibyl
\citep{wang-etal-2025-sibyl}. \\
\bottomrule
\end{tabular}
\caption{Baseline groups used in the main comparison.}
\label{tab:baseline-groups}
\end{table*}

The baselines differ in the additional control or training signal used on top
of the shared Mistral backbone. LLM-only uses the base instruction-tuned
generator without affective control; LLM-prompt uses verbal affective
instructions; LLM-SFT uses supervised response learning without the latent
affective-orientation module; EmPO-DPO uses preference optimization; and CASE,
APTNESS, and Sibyl are reproduced as ED-compatible Mistral-based variants of
their respective task-specific or commonsense-conditioning mechanisms.

The prior-method baselines are not evaluated using released outputs from the
original papers. Instead, they are reproduced within our aligned evaluation
pipeline to control for backbone, input format, test set, and decoding setup.
The results should therefore be interpreted as a controlled same-backbone
comparison of ED-compatible system variants rather than as an exact
replication of each prior system's original implementation, backbone, training
data, or compute environment. Where a baseline requires method-specific
training, conditioning, or auxiliary inference steps, we follow the
corresponding reproduced configuration and keep generation-time settings
matched whenever applicable.

\subsection{Human Evaluation Protocol}
\label{app:human-protocol}

Table~\ref{tab:human-evaluation-dimensions} defines the five dimensions used
in the main human evaluation.

\begin{table*}[t]
\centering
\small
\setlength{\tabcolsep}{3pt}
\begin{tabular}{p{0.16\textwidth}p{0.62\textwidth}p{0.11\textwidth}}
\toprule
Dimension & Evaluation question & Scoring direction \\
\midrule
Emotion appropriateness
& Which response better matches the emotional situation and the preceding
speaker's affective state?
& Positive \\

Felt responded
& Which response would make the previous speaker feel more seriously responded
to or understood?
& Positive \\

Context specificity
& Which response uses the concrete dialogue context more specifically rather
than giving a generic reply?
& Positive \\

Naturalness
& Which response sounds more natural as a human dialogue continuation?
& Positive \\

AI-like/problematic
& Which response sounds more template-like, excessive, didactic, or otherwise
problematic?
& Negative; reverse-scored \\
\bottomrule
\end{tabular}
\caption{
Human-evaluation dimensions. For positive dimensions, selecting
\textsc{EmoStance} is an \textsc{EmoStance} win. For the negative
AI-like/problematic dimension, selecting the baseline as more problematic is
counted as an \textsc{EmoStance} win.
}
\label{tab:human-evaluation-dimensions}
\end{table*}

The main blind pairwise evaluation was conducted in two batches with the same
instructions, blinding, and scoring procedure. The first batch used 10
annotators and 400 judgments. The second batch recruited 10 new annotators and
independently sampled new dialogue--response comparisons, adding another 400
judgments. The combined evaluation therefore contains 20 annotators, 40
judgments per annotator, and 800 judgments in total.

Each item contains one dialogue context, one evaluation question, and two
anonymized responses. Annotators do not see system names, emoji annotations,
latent affective-orientation regions, orientation vectors, or other
latent-control information. Response order is anonymized.

For positive dimensions, selecting the \textsc{EmoStance} response is counted
as a win. For AI-like/problematic phrasing, the response selected as more
problematic is counted as a loss for that system, so the dimension is
reverse-scored. Tie/Both equally good and Neither/Both bad are retained as
separate neutral categories and excluded from decisive win rates.

The analyses use individual judgments rather than treating an item-majority
label as a unique gold preference. We report 95\% Wilson confidence intervals
over decisive judgments and two-sided exact sign tests. The main per-baseline
table additionally applies Holm correction across the seven baseline
comparisons.

The evaluation is intentionally interpreted as preference evidence rather than
as recovery of a unique correct response. Fine-grained dialogue judgments are
subjective, and different annotators can reasonably prefer different plausible
affective orientations or response realizations.

\subsection{Dimension-Level Human Results}
\label{app:human-dimension-results}

Table~\ref{tab:human-dimension-results} groups the 800 blind pairwise
judgments by the evaluation question used for each item. Context specificity
and felt responded show the clearest gains. Emotion appropriateness and
naturalness have positive point estimates, but their confidence intervals
include parity. AI-like/problematic phrasing provides no evidence of
improvement.

\begin{table*}[t]
\centering
\small
\begin{tabular}{lrrrrr}
\toprule
Dimension & Win & Neutral & Lose & Win\% & 95\% CI \\
\midrule
Emotion appropriateness & 75  & 26 & 59 & 56.0 & [47.5, 64.1] \\
Felt responded          & 97  & 28 & 35 & 73.5 & [65.4, 80.3] \\
Context specificity     & 101 & 27 & 32 & 75.9 & [68.0, 82.4] \\
Naturalness             & 71  & 37 & 52 & 57.7 & [48.9, 66.1] \\
AI-like/problematic     & 51  & 47 & 62 & 45.1 & [36.3, 54.3] \\
\bottomrule
\end{tabular}
\caption{
Expanded human preference results by evaluation dimension. Neutral combines
Tie and Neither/Both bad. AI-like/problematic is reverse-scored because the
selected response is the more problematic one.
}
\label{tab:human-dimension-results}
\end{table*}

\subsection{Focused Human-Ablation Details}
\label{app:human-ablation-details}

The focused human ablation compares the final \textsc{EmoStance} system with
three deployable variants: without reranking, without the role-aware
response-orientation predictor, and without orientation control. As in the
main evaluation, we collected a second batch from 10 new annotators using
newly sampled contexts and the same blind pairwise protocol.

The combined study contains 20 annotators and 900 judgments. Each comparison
covers 100 dialogue contexts with three judgments per context, yielding 300
judgments per ablation.

At the judgment level, \textsc{EmoStance} receives 156 wins, 53 ties, 18
neither/both-bad outcomes, and 73 losses against the variant without
reranking. Excluding neutral outcomes, this corresponds to a 68.1\% decisive
win rate.

Against the variant without role-aware response-orientation prediction,
\textsc{EmoStance} receives 138 wins, 66 ties, 18 neither/both-bad outcomes,
and 78 losses, corresponding to a 63.9\% decisive win rate.

Against zero control, \textsc{EmoStance} receives 222 wins, 20 ties, 21
neither/both-bad outcomes, and 37 losses, corresponding to an 85.7\% decisive
win rate.

Across the three comparisons, the final system receives 516 wins, 139 ties, 57
neither/both-bad outcomes, and 188 losses. This gives a 73.3\% overall
decisive win rate. All three comparisons have two-sided exact sign-test values
below .001. Table~\ref{tab:human-ablation} in the main paper reports the
corresponding Wilson confidence intervals.

\subsection{Automatic Main Evaluation}
\label{app:auto-main}

Table~\ref{tab:automatic-main} presents system-level automatic results on the
full ED test set. We report reference-based similarity metrics and surface-form
diagnostics. BERTScore-F1 is the main semantic-similarity measure; ROUGE-L,
BLEU-2, and METEOR are included for comparability with prior work. Distinct-1/2,
Self-BLEU, and Generic measure diversity and template-like response rates.
These automatic metrics are treated as diagnostic indicators rather than
substitutes for human preference.

\begin{table*}[t]
\centering
\small
\setlength{\tabcolsep}{4pt}
\begin{tabular}{lrrrrrrrr}
\toprule
Method
& BERTScore \(\uparrow\)
& R-L \(\uparrow\)
& B-2 \(\uparrow\)
& METEOR \(\uparrow\)
& Dist-1 \(\uparrow\)
& Dist-2 \(\uparrow\)
& Self-BLEU \(\downarrow\)
& Generic \(\downarrow\) \\
\midrule
\textsc{EmoStance} / Ours
& 0.6523 & 0.1453 & 0.0399 & 0.1594
& 0.0416 & 0.2450 & 0.6804 & 0.4186 \\
LLM-only
& 0.6348 & 0.0908 & 0.0168 & 0.2042
& 0.0344 & 0.2947 & 0.6204 & 0.0228 \\
LLM-prompt
& 0.6353 & 0.0867 & 0.0152 & 0.2199
& 0.0272 & 0.2404 & 0.6792 & 0.0057 \\
LLM-SFT
& 0.6440 & 0.1310 & 0.0329 & 0.1624
& 0.0427 & 0.2624 & 0.6464 & 0.2634 \\
EmPO-DPO
& 0.6479 & 0.1305 & 0.0345 & 0.1694
& 0.0507 & 0.3489 & 0.5434 & 0.1412 \\
CASE
& 0.6240 & 0.1354 & 0.0309 & 0.1379
& 0.0072 & 0.0282 & 0.9315 & 0.4725 \\
APTNESS
& 0.6187 & 0.0768 & 0.0107 & 0.1906
& 0.0264 & 0.2292 & 0.6914 & 0.0021 \\
Sibyl
& 0.6492 & 0.1115 & 0.0243 & 0.2027
& 0.0375 & 0.2716 & 0.6502 & 0.1743 \\
\bottomrule
\end{tabular}
\caption{
System-level automatic evaluation on the full ED test set. Reference-based
metrics compare generated responses with ED references. Distinct-1/2 and
Self-BLEU measure diversity; Generic measures template-like responses.
}
\label{tab:automatic-main}
\end{table*}

The automatic comparison should be interpreted separately from human
preference. Reference-based metrics show that \textsc{EmoStance} is well
aligned with ED references under BERTScore-F1, ROUGE-L, and BLEU-2, but they
do not by themselves establish superior empathy or naturalness. The diversity
diagnostics also do not support a uniform diversity claim:
\textsc{EmoStance} is neither the most lexically diverse system nor the least
generic system.

\paragraph{Interpreting the Generic diagnostic.}
The Generic score should be interpreted as a surface-form diagnostic rather
than as a direct measure of context specificity. In our implementation, Generic
flags responses that match fixed generic-response rules and very short
responses with at most four tokens. It can therefore detect short or formulaic
realizations, but it cannot determine whether a response semantically takes up
the concrete dialogue context.

This distinction helps explain why \textsc{EmoStance} can have a relatively
high Generic rate in Table~\ref{tab:automatic-main} while still being
preferred by humans on context specificity and felt responded. The human
dimensions evaluate response quality at the dialogue-semantic level. At the
same time, the high Generic rate reflects a real surface-level limitation:
\textsc{EmoStance} sometimes realizes the predicted orientation through safe,
short, or formulaic empathetic phrasing. We therefore interpret
\textsc{EmoStance} as improving contextual uptake and perceived
responsiveness rather than as eliminating template-like surface phrasing.

\subsection{Automatic Component Ablations}
\label{app:auto-ablation}

We evaluate three functional components of \textsc{EmoStance}:
response-orientation prediction, continuous orientation-vector construction,
and generation-time control. Table~\ref{tab:ablation-design} summarizes the
ablation design.

\begin{table*}[t]
\centering
\small
\begin{tabular}{p{0.17\textwidth}p{0.24\textwidth}p{0.52\textwidth}}
\toprule
Component & Variant & Purpose \\
\midrule
Orientation prediction
& Context-only predictor
& Predicts source-side affective expression and response-side orientation from
text context only. \\

Orientation prediction
& Full role-aware predictor
& Adds speaker-role features, role-transition features, source-expression
conditioning, and a gated transition prior. \\

Orientation prediction
& Role-blind target prediction
& Removes role and transition features to test whether dyadic role structure
helps response-orientation prediction. \\

Orientation prediction
& Neural-only target prediction
& Removes the gated transition prior while retaining neural
response-orientation prediction. \\

Orientation prediction
& Hard-target supervision
& Replaces soft orientation distributions with argmax labels to test the value
of distributional supervision. \\

Vector construction
& Direct vector regression
& Predicts the 256-dimensional response-side control vector directly. \\

Vector construction
& Prototype reconstruction
& Predicts a latent-region distribution and reconstructs the vector as a
mixture of orientation prototypes. \\

Generation control
& Zero control
& Uses the prefix architecture with a null control vector. \\

Generation control
& Shuffled control
& Intentionally mismatches orientation vectors across examples as a negative
control. \\

Generation control
& Predicted control
& Uses the response orientation predicted from deployable text input. \\

Generation control
& Role-aware predicted control
& Uses the role-aware orientation predictor. \\

Generation control
& Role-aware control + reranking
& Samples multiple candidates and selects the response whose realized
orientation best matches the intended control. \\

Generation control
& Reference-conditioned controls
& Use reference-response orientation information and are reported only as
non-deployable upper-reference conditions. \\
\bottomrule
\end{tabular}
\caption{
Ablation design. Deployable variants use only inference-time text input.
Reference-conditioned variants use reference-response information and are not
deployable systems.
}
\label{tab:ablation-design}
\end{table*}

For orientation prediction, we report soft cross-entropy (CE),
Jensen--Shannon divergence (JSD), macro-F1, and Brier score. CE and JSD measure
distributional closeness to weak response-orientation targets; macro-F1
accounts for latent-region imbalance; and Brier score measures probability
quality. For continuous orientation vectors, we report target-vector cosine
similarity and mean squared error (MSE). For generation control, we report
orientation consistency against the weak response-derived target distribution,
together with generic-response and repetition rates. These automatic
generation-control metrics are diagnostics of control realization rather than
direct measures of human preference.

\subsubsection{Response-Orientation Prediction}

\begin{table*}[t]
\centering
\small
\begin{tabular}{lrrrr}
\toprule
Method & CE \(\downarrow\) & JSD \(\downarrow\)
& Macro-F1 \(\uparrow\) & Brier \(\downarrow\) \\
\midrule
Full role-aware predictor
& 1.3792 & 0.1823 & 0.3260 & 0.2578 \\
Context-only DeBERTa
& 1.3741 & 0.1706 & 0.3140 & 0.2546 \\
w/o role-aware transition
& \(1.3908\pm0.0061\) & \(0.1817\pm0.0012\)
& \(0.3198\pm0.0040\) & \(0.2609\pm0.0025\) \\
w/o gated transition prior
& \(1.3888\pm0.0080\) & \(0.1823\pm0.0013\)
& \(0.3231\pm0.0011\) & \(0.2610\pm0.0032\) \\
Hard-label training
& \(1.4450\pm0.0034\) & \(0.1851\pm0.0019\)
& \(0.3067\pm0.0062\) & \(0.2790\pm0.0008\) \\
\bottomrule
\end{tabular}
\caption{
Response-orientation prediction ablations. Newly run stochastic ablations are
reported as mean \(\pm\) standard deviation over seeds 13, 21, and 42. The
full and context-only rows are single-run results.
}
\label{tab:orientation-prediction}
\end{table*}

The full role-aware predictor is useful but not uniformly dominant across all
automatic metrics. It achieves the best macro-F1, whereas the context-only
predictor has slightly lower CE, JSD, and Brier score. The ablation rows are
more informative for the design choice: removing role-aware transition
information or removing the gated transition prior worsens CE relative to the
full predictor, while hard-target supervision substantially degrades CE and
macro-F1.

\subsubsection{Continuous Orientation-Vector Construction}

\begin{table}[t]
\centering
\small
\setlength{\tabcolsep}{3pt}
\begin{tabular}{lrr}
\toprule
Method
& \shortstack{Target cosine\\\(\uparrow\)}
& \shortstack{Target MSE\\\(\downarrow\)} \\
\midrule
Prototype reconstruction & 0.9236 & 0.000022 \\
Direct vector regression & 0.3220 & 0.001058 \\
\bottomrule
\end{tabular}
\caption{
Continuous target-vector construction. Prototype reconstruction maps the
predicted orientation distribution to a mixture of orientation prototypes,
while direct regression predicts the continuous vector directly.
}
\label{tab:orientation-vector-construction}
\end{table}

Prototype reconstruction produces a substantially more stable predicted
control vector than direct regression. The paired bootstrap comparison gives a
target-vector cosine delta of 0.6015 with a 95\% confidence interval of
[0.5997, 0.6033] and an MSE delta of \(-0.0010\) with a 95\% confidence
interval of \([-0.00105,-0.00103]\), favoring prototype reconstruction.
This comparison demonstrates predictability under the proposed supervision; it
does not establish lossless reconstruction. The deployable vector remains
constrained to mixtures of the nine prototypes and can omit
response-specific residual variation present in the dense weak target.

\subsubsection{Generation Control}

\begin{table*}[t]
\centering
\scriptsize
\setlength{\tabcolsep}{2.5pt}
\begin{tabular}{llrrrrr}
\toprule
Method & Status
& CE \(\downarrow\)
& JSD \(\downarrow\)
& \shortstack{Macro-F1\\\(\uparrow\)}
& \shortstack{Generic\\\(\downarrow\)}
& \shortstack{Repetition\\\(\downarrow\)} \\
\midrule

Zero control
& Diagnostic
& \(1.7816\pm0.0173\)
& \(0.2041\pm0.0025\)
& \(0.2584\pm0.0039\)
& \(0.0326\pm0.0023\)
& \(0.0020\pm0.0020\) \\

Shuffled control
& Diagnostic
& \(2.0570\pm0.0571\)
& \(0.2286\pm0.0092\)
& \(0.2332\pm0.0162\)
& \(0.1921\pm0.0098\)
& \(0.0111\pm0.0098\) \\

\shortstack[l]{Reference-conditioned\\control}
& Upper reference
& \(1.2678\pm0.0349\)
& \(0.1089\pm0.0066\)
& \(0.4162\pm0.0098\)
& \(0.1589\pm0.0181\)
& \(0.0111\pm0.0074\) \\

Predicted control
& Deployable
& \(1.7576\pm0.0572\)
& \(0.1791\pm0.0099\)
& \(0.2920\pm0.0183\)
& \(0.2708\pm0.0277\)
& \(0.0078\pm0.0020\) \\

\shortstack[l]{Role-aware predicted\\control}
& Deployable
& \(1.7367\pm0.0449\)
& \(0.1800\pm0.0080\)
& \(0.3040\pm0.0203\)
& \(0.2682\pm0.0274\)
& \(0.0078\pm0.0020\) \\

\shortstack[l]{Role-aware control\\+ rerank}
& Deployable
& \(\mathbf{1.5568\pm0.0299}\)
& \(\mathbf{0.1704\pm0.0048}\)
& \(\mathbf{0.3412\pm0.0117}\)
& \(\mathbf{0.1641\pm0.0119}\)
& \(\mathbf{0.0065\pm0.0041}\) \\

\shortstack[l]{Reference-conditioned\\selection}
& Upper reference
& \(1.2583\pm0.0277\)
& \(0.1141\pm0.0066\)
& \(0.4416\pm0.0453\)
& \(0.1465\pm0.0119\)
& \(0.0072\pm0.0011\) \\
\bottomrule
\end{tabular}
\caption{
Generation-control diagnostics over three seeds on 512-example test subsets.
The best deployable result is bolded. Reference-conditioned rows use
reference-response information and are not deployable. These metrics measure
orientation consistency and degeneration diagnostics rather than human
preference.
}
\label{tab:generation-control}
\end{table*}

Among deployable systems, role-aware predicted control with reranking obtains
the best orientation-consistency diagnostics: it has the lowest CE and JSD and
the highest macro-F1. It also reduces generic-response degeneration relative to
role-aware predicted control without reranking.

Reference-conditioned rows directly observe the orientation realized in the ED
reference response and are included only as non-deployable upper-reference
conditions. Their advantage over predicted control combines model error with
contextual underdetermination, because several response-side orientations may
be reasonable before the actual reference response is observed.

\subsection{Supplementary Ablation Diagnostics}
\label{app:supp-ablation}

This subsection collects secondary diagnostics that are useful for analysis but
are not the primary basis for the paper's claims. Top-1 accuracy collapses soft
orientation distributions into hard labels, expected calibration error depends
on binning choices, and generation-control diversity or control-realization
metrics are descriptive system analyses rather than direct measures of
empathy, naturalness, or human preference.

\subsubsection{Supplementary Orientation-Prediction Metrics}

\begin{table}[t]
\centering
\small
\setlength{\tabcolsep}{1.5pt}
\begin{tabular}{@{}lrr@{}}
\toprule
Method
& \shortstack{Accuracy\\\(\uparrow\)}
& \shortstack{ECE\\\(\downarrow\)} \\
\midrule

\shortstack[l]{Full role-aware\\predictor}
& 0.5316
& 0.0404 \\

\shortstack[l]{Context-only\\DeBERTa}
& 0.5428
& 0.0152 \\

\shortstack[l]{w/o role-aware\\transition}
& \(0.5266\pm0.0005\)
& \(0.0294\pm0.0066\) \\

\shortstack[l]{w/o gated\\transition prior}
& \(0.5287\pm0.0023\)
& \(0.0349\pm0.0095\) \\

\shortstack[l]{Hard-label\\training}
& \(0.5286\pm0.0016\)
& \(0.0311\pm0.0109\) \\

Graph-only
& 0.4846
& 0.0581 \\

\shortstack[l]{Calibrated graph\\fusion}
& 0.5428
& 0.0152 \\

\bottomrule
\end{tabular}
\caption{
Supplementary orientation-prediction metrics. Accuracy is less central than CE
and JSD because the weak target is a soft distribution rather than a hard gold
label.
}
\label{tab:orientation-supplementary}
\end{table}

Accuracy measures whether the most probable predicted region matches the most
probable weak target region. ECE measures confidence calibration after binning
predicted probabilities. The graph-only diagnostic is weaker than text-based
prediction, indicating that the induced emoji/orientation graph should not be
treated as a stand-alone response-orientation classifier. In the available
calibrated-fusion artifact, the selected fusion weight is zero, so calibrated
graph fusion matches the context-only predictor. These supplementary metrics
reinforce the caution used in the main paper: the role-aware predictor is not
uniformly best across all intrinsic metrics.

\subsubsection{Source-Vector Feature Ablation}

\begin{table*}[t]
\centering
\small
\begin{tabular}{lrrrr}
\toprule
Method
& Target CE \(\downarrow\)
& Macro-F1 \(\uparrow\)
& Target cosine \(\uparrow\)
& Source cosine \(\uparrow\) \\
\midrule
No source-vector feature
& 1.3840 & 0.3125 & 0.9225 & 0.2120 \\
Direct source-vector feature
& 1.3794 & 0.3253 & 0.9230 & 0.2445 \\
Prototype source-vector feature
& 1.3794 & 0.3224 & 0.9229 & 0.9431 \\
\bottomrule
\end{tabular}
\caption{
Supplementary source-vector feature ablation. Source-vector cosine compares
feature stability, while target CE and macro-F1 indicate the effect on
response-orientation prediction.
}
\label{tab:source-vector-ablation}
\end{table*}

These source-vector results are diagnostic rather than central to the main
claims. They indicate that source-side affective-expression features can
influence response-orientation prediction, but they do not replace the main
evidence that prototype reconstruction provides a stable mechanism for
constructing the response-side control vector used by the generator.

\subsubsection{Supplementary Generation-Control Diversity Diagnostics}

\begin{table*}[t]
\centering
\small
\begin{tabular}{llrrr}
\toprule
Method & Status
& Accuracy \(\uparrow\)
& Distinct-2 \(\uparrow\)
& Self-BLEU \(\downarrow\) \\
\midrule
Zero control
& Diagnostic
& \(0.4837\pm0.0163\)
& \(0.5360\pm0.0308\)
& \(0.6863\pm0.0173\) \\

Shuffled control
& Diagnostic
& \(0.4368\pm0.0236\)
& \(0.5125\pm0.0150\)
& \(0.5663\pm0.0128\) \\

Reference-conditioned control
& Upper reference
& \(0.6348\pm0.0141\)
& \(0.5270\pm0.0081\)
& \(0.5752\pm0.0133\) \\

Predicted control
& Deployable
& \(0.5182\pm0.0262\)
& \(0.4875\pm0.0139\)
& \(0.5901\pm0.0136\) \\

Role-aware predicted control
& Deployable
& \(0.5169\pm0.0283\)
& \(0.4948\pm0.0214\)
& \(0.5803\pm0.0166\) \\

Role-aware control + rerank
& Deployable
& \(\mathbf{0.5365\pm0.0354}\)
& \(\mathbf{0.5803\pm0.0086}\)
& \(\mathbf{0.4689\pm0.0017}\) \\

Reference-conditioned selection
& Upper reference
& \(0.6556\pm0.0229\)
& \(0.5715\pm0.0143\)
& \(0.4792\pm0.0123\) \\
\bottomrule
\end{tabular}
\caption{
Supplementary generation-control diversity diagnostics. The best deployable
result is bolded.
}
\label{tab:generation-diversity}
\end{table*}

Distinct-2 and Self-BLEU describe response diversity and are not interpreted as
direct human-preference metrics. The reranked deployable system obtains the
best supplementary accuracy, Distinct-2, and Self-BLEU values among deployable
variants in this diagnostic setting. These metrics describe surface-form and
internal orientation-scoring behavior only.

\subsubsection{Intended-Control Realization Diagnostics}

\begin{table*}[t]
\centering
\scriptsize
\setlength{\tabcolsep}{4pt}
\begin{tabular}{llrrrr}
\toprule
Method & Status
& Intended CE \(\downarrow\)
& Intended JSD \(\downarrow\)
& Intended Acc. \(\uparrow\)
& Intended Macro-F1 \(\uparrow\) \\
\midrule
Zero control
& Diagnostic
& -- & -- & -- & -- \\

Shuffled control
& Diagnostic
& \(2.1077\pm0.0636\)
& \(0.2302\pm0.0122\)
& \(0.4284\pm0.0023\)
& \(0.2402\pm0.0136\) \\

Reference-conditioned control
& Upper reference
& \(1.2678\pm0.0349\)
& \(0.1089\pm0.0066\)
& \(0.6348\pm0.0141\)
& \(0.4162\pm0.0098\) \\

Predicted control
& Deployable
& \(1.7018\pm0.0066\)
& \(0.0762\pm0.0018\)
& \(0.6510\pm0.0149\)
& \(0.4025\pm0.0236\) \\

Role-aware predicted control
& Deployable
& \(1.9668\pm0.0172\)
& \(0.0922\pm0.0021\)
& \(0.6380\pm0.0181\)
& \(0.4197\pm0.0413\) \\

Role-aware control + rerank
& Deployable
& \(1.6708\pm0.0120\)
& \(0.0535\pm0.0013\)
& \(0.7031\pm0.0119\)
& \(0.6285\pm0.0415\) \\

Reference-conditioned selection
& Upper reference
& \(1.2583\pm0.0277\)
& \(0.1141\pm0.0066\)
& \(0.6556\pm0.0229\)
& \(0.4416\pm0.0453\) \\
\bottomrule
\end{tabular}
\caption{
Control-realization diagnostics against the intended orientation distribution.
These metrics evaluate whether the generator realizes the supplied control,
not whether the response is preferred by humans.
}
\label{tab:intended-control}
\end{table*}

The zero-control condition has no intended orientation distribution and is
therefore marked as unavailable. These diagnostics show that reranking improves
realization of the supplied control among deployable systems. The result is
useful for validating the control mechanism, but it is not itself a
human-quality result.

\subsubsection{Targeted Bootstrap Diagnostics for Generation Control}

\begin{table*}[t]
\centering
\small
\begin{tabular}{llll}
\toprule
System A & System B & Metric & Delta [95\% CI] \\
\midrule
Role-aware + rerank
& Role-aware
& Target CE \(\downarrow\)
& \(-0.1799\) [\(-0.2240,-0.1349\)] \\

Role-aware + rerank
& Role-aware
& Target JSD \(\downarrow\)
& \(-0.0097\) [\(-0.0157,-0.0035\)] \\

Role-aware + rerank
& Role-aware
& Generic \(\downarrow\)
& \(-0.1042\) [\(-0.1263,-0.0827\)] \\
\bottomrule
\end{tabular}
\caption{
Targeted context-level paired-bootstrap diagnostics for generation-control
ablation. Deltas are computed as System A minus System B. These comparisons
evaluate automatic orientation consistency and degeneration behavior rather
than human preference.
}
\label{tab:control-bootstrap}
\end{table*}

The paired-bootstrap comparisons support the diagnostic claim that reranking
improves automatic orientation consistency and reduces generic-response
behavior relative to the non-reranked role-aware variant. Human preference is
assessed separately through the focused ablation evaluation.
\section{Reproducibility, Artifacts, and Human-Participant Details}
\label{app:reproducibility}

This appendix provides checklist-related reproducibility information for
artifacts, licenses and intended use, privacy, compute, hyperparameters,
software, human-participant procedures, and AI-assistance disclosure.
The accompanying code artifact contains the full implementation,
configuration files, preprocessing scripts, and evaluation scripts.

\subsection{Artifacts, Licenses, and Intended Use}
\label{app:artifact_licenses}

Table~\ref{tab:artifact_inventory} summarizes the main artifacts used or
created in this work. Existing datasets, pretrained models, software
packages, and baseline outputs are used for research on empathetic
dialogue generation. The derived emoji annotations and latent stance
representations are intended as weak supervision for research, not as
gold emotion labels, user-profiling signals, clinical indicators, or
mental-state diagnoses.

If released, EmojiDialogue will be distributed only under terms
compatible with the original EmpatheticDialogues license. We will release
annotation metadata and construction scripts, such as example
identifiers, emoji annotations, confidence scores, soft emoji
distributions, induced stance-cluster assignments, and preprocessing
code. We will not redistribute the original EmpatheticDialogues dialogue
text or situation descriptions. Users who wish to reconstruct the full
resource should obtain EmpatheticDialogues under its own access and
license conditions and then apply our released metadata and scripts.
\begin{table*}[t]
\centering
\scriptsize
\setlength{\tabcolsep}{3pt}
\renewcommand{\arraystretch}{1.08}
\begin{tabular}{
@{}
p{0.15\textwidth}
p{0.22\textwidth}
p{0.39\textwidth}
p{0.17\textwidth}
@{}
}
\toprule
Artifact
& Version / source
& Use in this paper
& License / terms
\\
\midrule

EmpatheticDialogues
&
Official EmpatheticDialogues split
&
Source English dyadic dialogue corpus for constructing
\textsc{EmojiDialogue} and for EmpatheticDialogues response-generation
evaluation.
&
CC BY-NC 4.0; non-commercial research use.
\\

Derived \textsc{EmojiDialogue} metadata
&
Constructed in this work
&
Emoji annotations, confidence scores, soft emoji distributions, latent
stance clusters, continuous stance vectors, and construction metadata.
&
Research-only, non-commercial, CC BY-NC 4.0-compatible terms. The
accompanying artifact contains metadata and scripts only, not original
dialogue text.
\\

Code artifact and scripts
&
Accompanying code artifact
&
Preprocessing, stance construction, training, decoding, reranking, and
evaluation scripts.
&
MIT License.
\\

Python emoji package
&
\texttt{emoji==0.1.0}
&
Initial emoji universe construction before human screening.
&
BSD License.
\\

LLM annotators
&
DeepSeek-V3.2; Claude-Sonnet-4.6; Gemini-2.5-Pro; GPT-5.4, accessed via
API in March--April 2026
&
Training-time weak emoji annotation only. These models are not used at
\textsc{EmoStance} inference time.
&
Provider terms of service; model weights are not redistributed.
\\

Frozen generator and baseline base model
&
\texttt{mistralai/}\allowbreak
\texttt{Mistral-7B-Instruct-v0.3}
&
Frozen generator for \textsc{EmoStance} and base generator for reproduced
baselines.
&
Apache-2.0.
\\

Context / stance encoder
&
\texttt{microsoft/}\allowbreak
\texttt{deberta-v3-base}
&
Encoder used by the DeBERTa-based stance modules.
&
MIT.
\\

Utterance representation / clustering pipeline
&
In-repository hashed TF--IDF utterance encoder and emoji-centroid
pipeline
&
Computes \(\psi(u_t)\) for name-free clustering, emoji centroids, and
stance-vector construction. This is not a DeBERTa checkpoint.
&
Covered by the accompanying MIT-licensed code artifact.
\\

\bottomrule
\end{tabular}
\caption{Artifacts, sources, licenses, and intended uses.}
\label{tab:artifact_inventory}
\end{table*}
All baseline outputs used in the main comparison are produced by our own
reproduction under the aligned EmpatheticDialogues evaluation setting
using \texttt{mistralai/Mistral-7B-Instruct-v0.3} as the base generator.
The reproduced outputs are used for research comparison and are subject
to the same dataset-use restrictions as EmoStance outputs.

The intended use of EmojiDialogue and EmoStance is research on weakly
supervised listener-stance modeling and empathetic response generation.
The derived annotation layer should not be used as gold emotion
annotation, psychological diagnosis, protected-attribute inference,
user profiling, clinical decision making, or evidence of a user's true
internal mental state.

\subsection{Privacy, Identifying Information, and Sensitive Content}
\label{app:privacy_checks}

We do not collect new dialogue data from speakers. The dialogue text
comes from the publicly released EmpatheticDialogues benchmark. Our
added annotation layer consists of emoji labels, confidence scores, soft
emoji distributions, latent stance clusters, continuous stance vectors,
and derived stance-control representations. The annotation and
stance-construction pipeline does not add names, usernames, email
addresses, phone numbers, locations, account identifiers, or other
direct personal identifiers.

We do not infer protected attributes such as gender, ethnicity, health
status, political views, sexual orientation, disability status, or other
sensitive demographic properties. The emoji labels and latent stance
representations are treated as weak conversational stance signals rather
than as evidence of a speaker's true internal state, identity, or
demographic attributes.

We did not conduct a separate exhaustive PII audit beyond using the
publicly released benchmark and ensuring that our annotation pipeline
does not add new personal identifiers. This means that we cannot
guarantee that the original benchmark contains no residual identifying
information. To reduce redistribution risk, any future release of
EmojiDialogue will avoid redistributing the original text and will
release only annotation metadata and construction scripts, consistent
with Appendix~\ref{app:artifact_licenses}.

Human annotator identities are not linked to released dialogue examples.
Human-evaluation results are reported only in aggregate. We do not
release individual annotator identities together with item-level
judgments. If item-level annotation metadata are released, they will not
include annotator names, contact information, raw API logs, provider
account metadata, timestamps, or other information that could link
individual annotators or API accounts to specific judgments.

We did not perform additional offensive-content filtering beyond the
original benchmark preprocessing, because emotionally grounded dialogue
may naturally include distressing, sensitive, or personally framed
experiences. Human annotation and evaluation results should therefore be
interpreted as research judgments over benchmark dialogue, not as
judgments about real users or clinical cases.

\subsection{Model Size, Infrastructure, and Compute Budget}
\label{app:model-compute}

Table~\ref{tab:model-compute} summarizes the model sizes, trainable
components, runtime environment, and approximate training budget. The frozen
generator is instantiated as
\texttt{mistralai/}\allowbreak\texttt{Mistral-7B-Instruct-v0.3} and is not
updated during \textsc{EmoStance} training. The context and orientation modules use
\texttt{microsoft/deberta-v3-base}, a
\mbox{DeBERTaV3} encoder
\citep{he-etal-2023-debertav3}. The trainable
components consist of the DeBERTa-based orientation modules and the prefix
projector.

The utterance representation \(\psi(u_t)\) used for name-free clustering,
emoji-centroid construction, and orientation-vector construction is produced
by the in-repository hashed TF--IDF and emoji-centroid pipeline rather than by
the DeBERTa checkpoint. Full model definitions and trainable-parameter details
are provided in the supplementary code artifact.

\begin{table}[t]
\centering
\small
\setlength{\tabcolsep}{4pt}
\begin{tabular}{p{0.36\columnwidth}p{0.56\columnwidth}}
\toprule
Item & Value \\
\midrule

Frozen generator
&
\texttt{mistralai/}\allowbreak
\texttt{Mistral-7B-Instruct-v0.3}
\\

Generator size
&
Approximately 7B parameters
\\

Context / orientation encoder
&
\texttt{microsoft/}\allowbreak
\texttt{deberta-v3-base}
\\

Encoder size
&
Approximately 184M parameters
\\

Utterance representation
&
In-repository hashed TF--IDF / emoji-centroid pipeline
\\

Trainable modules
&
DeBERTa-based orientation modules and prefix projector
\\

Trainable-parameter details
&
Provided in the supplementary code artifact
\\

Hardware
&
\(1\times\) NVIDIA RTX 4090
\\

Precision
&
bf16 for the frozen Mistral generator; fp32 for the DeBERTa-based
orientation modules
\\

Main \textsc{EmoStance} training time
&
Approximately 2--3 wall-clock hours
\\

Main \textsc{EmoStance} training budget
&
Approximately 2--3 GPU-hours
\\

Additional decoding, ablation, and baseline runs
&
Same single-GPU environment; not separately logged
\\

Efficiency-oriented decoding
&
\(B=1\); single controlled generation without reranking
\\

Quality-oriented decoding
&
\(B=4\); orientation-consistency reranking
\\

\bottomrule
\end{tabular}
\caption{
Model size, infrastructure, and approximate compute budget. GPU-hour values
are approximate single-GPU estimates. Decoding, reranking, ablation, baseline,
and API-annotation costs are not included in the main training budget.
}
\label{tab:model-compute}
\end{table}

GPU-hour estimates are computed as wall-clock hours multiplied by the number
of GPUs. Because the experiments use a single GPU, GPU-hours equal wall-clock
hours for the reported main training run. These values are author-reported
running-environment estimates, and no separate GPU-hour log was retained for
the additional decoding, ablation, and baseline runs. The LLM annotators used
to construct the weak emoji annotations were accessed through external APIs
and are not included in the local GPU-hour budget. The inference cost of
multi-candidate reranking is also separate from the 2--3 GPU-hour main
training estimate and is quantified below.

\subsubsection{Inference-Efficiency Benchmark}
\label{app:inference-efficiency}

We benchmark end-to-end inference on a single NVIDIA RTX 4090 over a fixed
evaluation set under identical decoding settings. The efficiency-oriented
configuration generates one controlled response and performs no reranking
(\(B=1\)). The quality-oriented configuration generates four candidates,
scores each candidate for orientation consistency, and selects the
highest-scoring response (\(B=4\)). The two configurations otherwise use the
same model checkpoints and generation settings.

\begin{table}[!t]
\centering
\small
\setlength{\tabcolsep}{4pt}
\begin{tabular}{@{}lrr@{}}
\toprule
Metric
& No reranking
& Reranking
\\
\midrule

Candidates
& 1
& 4
\\

Mean latency
& 331.7 ms
& 1,333.4 ms
\\

P50 / P95 latency
& 317.4 / 548.3 ms
& 1,279.6 / 2,026.9 ms
\\

Throughput
& 3.015 ex./s
& 0.750 ex./s
\\

Relative cost
& \(1.00\times\)
& \(4.02\times\)
\\

\bottomrule
\end{tabular}
\caption{
End-to-end inference efficiency for single-generation decoding
and four-candidate orientation-consistency reranking. Both
configurations are measured on the same hardware under identical
decoding settings.
}
\label{tab:inference-efficiency}
\end{table}

\begin{table}[t]
\centering
\small
\setlength{\tabcolsep}{2pt}
\begin{tabular}{@{}p{0.44\columnwidth}rr@{}}
\toprule
Component
& \shortstack{Total\\time}
& \shortstack{Share of\\\(B=4\) time}
\\
\midrule

No-reranking pipeline
& 169.831 s
& --
\\

Full reranking pipeline
& 682.725 s
& 100.00\%
\\

\shortstack[l]{Four-candidate\\generation}
& 679.191 s
& 99.48\%
\\

Orientation scoring
& 3.268 s
& 0.48\%
\\

Final selection
& 0.267 s
& 0.04\%
\\

\bottomrule
\end{tabular}
\caption{
Component-level runtime profile of the four-candidate
orientation-consistency reranking pipeline.
}
\label{tab:inference-profile}
\end{table}

The \(B=4\) configuration increases mean end-to-end latency from approximately
0.332 seconds to 1.333 seconds per example, corresponding to a
\(4.02\times\) relative cost increase. Throughput decreases from 3.015 to
0.750 examples per second. Component-level profiling indicates that this
additional cost arises almost entirely from generating the extra candidates.

Orientation scoring and final selection together account for only 0.52\% of
the quality-oriented runtime, while four-candidate generation accounts for
99.48\%. The computational overhead therefore scales primarily with the number
of generated candidates rather than with the orientation scorer or final
selection operation.

This efficiency cost should be considered together with the expanded human
ablation in Table~\ref{tab:human-ablation}, where the reranked system achieves
a 68.1\% decisive win rate over the no-reranking configuration
(\(p<.001\)). We therefore present \(B=1\) as the lower-cost,
single-generation deployment mode and \(B=4\) as the quality-oriented mode
that trades approximately fourfold inference cost for higher human preference
and stronger orientation consistency.

The reported latency and throughput values are specific to the stated
hardware, implementation, prompt and response lengths, batch configuration,
and decoding settings. They should therefore be interpreted as a controlled
within-system comparison rather than as universal deployment figures.
\subsection{Hyperparameters, Model Selection, and Reporting Protocol}
\label{app:hyperparameters}

Hyperparameters were selected using the validation split. We did not
tune hyperparameters on the test set. Full configuration files are
included in the accompanying code artifact; Table~\ref{tab:hyperparameters}
reports the key values needed to interpret the main experiments.

\begin{table*}[t]
\centering
\scriptsize
\setlength{\tabcolsep}{4pt}
\renewcommand{\arraystretch}{1.08}
\begin{tabular}{
@{}
p{0.31\textwidth}
p{0.63\textwidth}
@{}
}
\toprule
Hyperparameter & Value \\
\midrule

Optimizer
&
AdamW
\\

Planner / stance-module learning rate
&
\(1.5\times10^{-5}\)
\\

Prefix-projector learning rate
&
\(1\times10^{-4}\)
\\

Planner batch size / effective batch size
&
8 per device / 8 effective
\\

Generator batch size / effective batch size
&
1 per device / 1 effective
\\

Gradient accumulation
&
Not used; 1
\\

Planner epochs
&
3
\\

Prefix-projector epochs
&
1
\\

Warmup
&
0.06 warmup ratio for the planner; not used for prefix-projector
training
\\

Weight decay
&
0.01 for the planner; the prefix projector uses the PyTorch AdamW
default and is not separately configured
\\

Maximum stance-input length
&
320 tokens
\\

Maximum generator prompt length
&
384 tokens
\\

Maximum response length during generator training
&
128 tokens
\\

Maximum new tokens during generation / reranking
&
64 tokens
\\

Latent stance clusters \(K\)
&
9
\\

Continuous stance dimension
&
256
\\

Prefix length \(m\)
&
8
\\

Projector hidden dimension
&
4096
\\

Role embedding dimension
&
32
\\

Main loss weights
&
\(\lambda_{\mathrm{tar}}=1.0\),
\(\lambda_{0}=0.2\),
\(\lambda_{\mathrm{src}}=0.4\),
\(\lambda_{\mathrm{vec}}=0.1\), and
\(\lambda_{\mathrm{tr}}=0.5\)
\\

Transition / imbalance settings
&
Transition smoothing \(\alpha=0.05\); focal \(\gamma=0.0\);
class-imbalance exponent \(\beta=0.25\)
\\

Emoji graph / clustering
&
\(\lambda_{\mathrm{ctx}}=0.65\),
\(\lambda_{\mathrm{conf}}=0.35\),
top-\(k=8\), centroid shrinkage \(\tau=50.0\);
Leiden resolution \(=1.6\)
\\

Boundary membership
&
Absolute threshold 0.70; relative threshold 0.85; maximum 4 clusters
\\

Soft-membership sharpening temperature
&
0.7
\\

Decoding settings
&
Sampling; temperature \(=0.7\), top-\(p=0.9\), top-\(k\) not used,
maximum new tokens \(=64\)
\\

Reranking
&
\(B=4\) candidates per input; \(\eta=0.0\) length penalty
\\

Random seeds
&
13, 21, and 42 for newly run stochastic ablations
\\

Main-result reporting
&
Role-aware stance checkpoint selected by development-set target soft
cross-entropy; generation-control and reranking diagnostics are reported
as mean \(\pm\) standard deviation over seeds 13, 21, and 42 on
512-example development/test subsets. The prefix projector is trained
for one epoch and the final checkpoint is used.
\\

\bottomrule
\end{tabular}
\caption{
Key hyperparameters. Full configuration files are provided in the
accompanying code artifact.
}
\label{tab:hyperparameters}
\end{table*}

The coefficient $\lambda_{0}$ corresponds to the prior-free target
auxiliary loss. The coefficient $\lambda_{\mathrm{tr}}$ controls the
strength of the transition-prior logits rather than an additive training
loss. Unless otherwise stated, rows explicitly described as single-run
artifacts are not averaged over seeds. Bootstrap confidence intervals,
Wilson confidence intervals, and two-sided sign tests are reported where
specified in the main text and appendices.

\subsection{Software and Metric Implementations}
\label{app:software_metrics}

Exact software versions and evaluation scripts are included in the
accompanying code artifact. Table~\ref{tab:software_metrics} records the
runtime environment and metric implementations most relevant for
reproducing the reported scores. Metric names follow the main automatic
evaluation tables.

\begin{table*}[!t]
\centering
\small
\begin{tabular}{p{0.25\linewidth} p{0.65\linewidth}}
\toprule
Component & Implementation / settings \\
\midrule
Runtime &
Python 3.10.14; PyTorch 2.6.0+cu118; CUDA 11.8 via
\texttt{torch.version.cuda} \\

Transformer stack &
\texttt{transformers==4.46.3}; \texttt{tokenizers==0.20.3};
\texttt{accelerate==1.13.0} \\

Data / numerical packages &
\texttt{datasets==2.19.1};
\texttt{numpy==2.2.6};
\texttt{pandas==2.3.3};
\texttt{scikit-learn==1.7.2};
\texttt{scipy==1.15.3};
\texttt{networkx==3.4.2} \\

PEFT / DeepSpeed &
\texttt{peft==0.19.1} installed but not used by the main EmoStance modules;
DeepSpeed not used \\

Emoji inventory &
\texttt{emoji==0.1.0} \\

Graph / clustering scripts &
In-repo hashed TF-IDF and emoji-centroid scripts; released clustering
artifacts are provided with the code artifact \\

BERTScore &
\texttt{bert-score==0.3.12}; model
\texttt{distilbert-base-uncased}; 6 layers; English \\

ROUGE-L &
Custom script; mean sentence-level LCS F1 with the project tokenizer \\

BLEU-2 &
Custom script; mean sentence-level BLEU-2 with clipped n-gram precision
and $10^{-9}$ precision-floor smoothing \\

METEOR &
Standard NLTK METEOR. \\

Distinct-1/2 &
Custom script; corpus-level unique unigram and bigram ratios over
generated responses \\

Self-BLEU &
Custom script; sentence BLEU-2 against other generated responses, with a
fixed-seed sample capped at 200 items. \\

Generic-response rate &
Custom rule-based diagnostic released with the code artifact; flags
responses matching fixed generic-response rules and very short responses
with at most four tokens as generic \\

\bottomrule
\end{tabular}
\caption{Software and metric implementations. Metric names follow the
main automatic evaluation tables.}
\label{tab:software_metrics}
\end{table*}

Automatic metrics are treated as diagnostics rather than substitutes for
human preference. Reference-based metrics measure similarity to ED
references, while diversity and generic-response diagnostics describe
surface-form behavior.
\subsection{Human Annotators and Participant Procedures}
\label{app:human-participants}

Human participants were involved in five procedures: emoji-inventory
screening, weak-annotation plausibility auditing, the human--LLM
emoji-distribution audit, the main blind pairwise evaluation, and the focused
ablation evaluation.

The emoji-inventory screening used 3 volunteer screeners. The plausibility
audit used 300 model--dialogue packages and 1,297 turn-level items, with three
annotations per turn. The human--LLM distributional audit used 3 human
annotators, 120 utterances, and 360 emoji judgments. The expanded main
pairwise evaluation used 20 annotators and 800 judgments. The expanded focused
ablation used 20 annotators, 100 dialogue contexts per comparison, three
judgments per context, and 900 judgments in total.

\paragraph{Recruitment and compensation.}
Annotators were recruited from lab members, graduate students, and university
student volunteers. They were unpaid volunteers. Participation was voluntary,
and annotators could withdraw at any time.

\paragraph{Consent and annotator information.}
Annotators were informed that their judgments would be used for research and
reported only in aggregate. They were informed that the task involved
emotionally grounded dialogue and could include sensitive or distressing
content. Annotators could skip items or withdraw from the study.

Annotators were fluent English speakers with NLP or dialogue-evaluation
background. The annotator pool was drawn from a China-based university
community. We did not collect individual-level recruitment-group labels beyond
the recruitment sources stated above, and we did not collect sensitive
demographic attributes such as gender, ethnicity, health status, political
views, sexual orientation, or disability status.

\paragraph{Ethics-review status.}
No formal ethics review was sought. The study involved aggregate evaluation of
publicly released benchmark dialogue and did not collect new dialogue data from
speakers. Annotators were informed of the task purpose, participated
voluntarily, and results are reported only in aggregate.

\paragraph{Emoji-inventory screening instructions.}
Screeners were asked whether each emoji could plausibly express an affective
state, interpersonal stance, or conversational attitude in dialogue. They were
instructed to include boundary cases when an emoji could reasonably convey
affect, orientation, attitude, hesitation, sympathy, celebration, concern,
embarrassment, teasing, or related conversational meanings.

\paragraph{Plausibility-audit instructions.}
For the weak-annotation plausibility audit, annotators saw a full dialogue
context, one utterance, and one emoji assigned by a hidden LLM annotator. They
judged whether the emoji plausibly expressed the utterance's affective state,
interpersonal stance, or conversational attitude in context. The available
labels were reasonable, questionable but acceptable, and clearly unreasonable.
Annotators were instructed that the goal was not to identify a gold-standard
emotion label or the speaker's true mental state, but only to judge contextual
plausibility.

\paragraph{Human--LLM distributional-audit instructions.}
Annotators saw the situation description, preceding dialogue context, current
speaker role, current utterance, and the same 136-emoji candidate inventory
used for LLM annotation. They independently selected exactly one emoji and
provided a 1--5 confidence score. LLM identities, LLM choices, LLM confidence
scores, and latent-region assignments were hidden. Annotators were told that
several emoji could be plausible and that the study would compare aggregated
distributions rather than treat any individual selection as a unique gold
label.

\paragraph{Main pairwise-evaluation instructions.}
Annotators saw a dialogue context and two anonymized candidate responses,
Response A and Response B. System names were hidden. Annotators answered one
evaluation question per item and selected A, B, Tie/Both equally good, or
Neither/Both bad. The five evaluation dimensions were emotion appropriateness,
felt responded, context specificity, naturalness, and AI-like/problematic
phrasing. The AI-like/problematic dimension was reverse-scored.

\paragraph{Focused-ablation instructions.}
Annotators saw a dialogue context and two anonymized responses. They answered
the following question: Which response better fits the dialogue context and
would make the previous speaker feel more seriously responded to or
understood? The available options were A, B, Tie/Both equally good, and
Neither/Both bad.
\subsection{AI Assistance Disclosure}
\label{app:ai_assistance}

We used LLMs as annotation tools to produce weak emoji labels, as
described in Section~3 and Appendix~\ref{app:llm-annotation-prompt}. These LLM annotations are part
of the experimental design and are treated as weak supervision rather
than gold labels. The LLM annotators are not used by EmoStance at inference
time.

AI assistants were used for limited language polishing, checklist
documentation, and code-editing assistance. All AI-assisted code edits
were reviewed, tested, and modified by the authors before use. All
scientific claims, experimental design choices, code, analyses, results,
and conclusions were reviewed and verified by the authors. AI assistants
were not credited as authors.

\end{document}